\documentclass[runningheads]{llncs}

\usepackage{tikz}
\usepackage{comment}
\usepackage{amsmath,amssymb} 
\usepackage{color}
\usepackage{graphicx}
\usepackage{amsmath}
\usepackage{soul}
\usepackage{textcomp}  
\usepackage{pifont}
\usepackage{enumitem}
\usepackage{colortbl}

\usepackage{graphicx}
\usepackage{booktabs}
\usepackage{rotating}

\usepackage{multirow}
\usepackage{gensymb}
\usepackage{subcaption}
\usepackage{lipsum}
\usepackage{blindtext}
\usepackage{xcolor}
\usepackage{subcaption}
\usepackage{breakcites}

\usepackage[accsupp]{axessibility}  

\usepackage{xspace}
\makeatletter
\DeclareRobustCommand\onedot{\futurelet\@let@token\@onedot}
\def\@onedot{\ifx\@let@token.\else.\null\fi\xspace}
\def\eg{\emph{e.g}\onedot} 
\def\ie{\emph{i.e}\onedot}

\def\etal{\emph{et al}\onedot}
\makeatother

\usepackage[pagebackref,breaklinks,colorlinks,
citecolor=blue,linkcolor=blue,urlcolor=blue]{hyperref}
\usepackage{orcidlink}
\usepackage[accsupp]{axessibility} 
\usepackage[width=122mm,left=12mm,paperwidth=146mm,height=193mm,top=12mm,paperheight=217mm]{geometry}

\begin{document}

\title{V2X-Real: a Large-Scale Dataset for Vehicle-to-Everything Cooperative Perception} 

\titlerunning{ }

\author{Hao Xiang, Zhaoliang Zheng, Xin Xia, Runsheng Xu, Letian Gao, Zewei Zhou, Xu Han, Xinkai Ji, Mingxi Li, Zonglin Meng, Li Jin, Mingyue Lei, Zhaoyang Ma, Zihang He, Haoxuan Ma, Yunshuang Yuan, Yingqian Zhao, Jiaqi Ma\thanks{Corresponding author. Work done at UCLA Mobility Lab.}}

\authorrunning{ }

\institute{University of California, Los Angeles}

\maketitle

\begin{abstract}
  Recent advancements in Vehicle-to-Everything (V2X) technologies have enabled autonomous vehicles to share sensing information to see through occlusions, greatly boosting the perception capability. However, there are no real-world datasets to facilitate the real V2X cooperative perception research -- existing datasets either only support Vehicle-to-Infrastructure cooperation or Vehicle-to-Vehicle cooperation. In this paper, we present V2X-Real, a large-scale dataset that includes a mixture of multiple vehicles and smart infrastructure to facilitate the V2X cooperative perception development with multi-modality sensing data. Our V2X-Real is collected using two connected automated vehicles and two smart infrastructure, which are all equipped with multi-modal sensors including LiDAR sensors and multi-view cameras. The whole dataset contains 33K LiDAR frames and 171K camera data with over 1.2M annotated bounding boxes of 10 categories in very challenging urban scenarios. According to the collaboration mode and ego perspective, we derive four types of datasets for Vehicle-Centric, Infrastructure-Centric, Vehicle-to-Vehicle, and Infrastructure-to-Infrastructure cooperative perception. Comprehensive multi-class multi-agent benchmarks of SOTA cooperative perception methods are provided. The V2X-Real dataset and codebase are available at \url{https://mobility-lab.seas.ucla.edu/v2x-real}. 
\end{abstract}

\section{Introduction}
\label{sec:intro}
The evolution of autonomous driving technology has been significantly accelerated by advancements in deep learning, 
particularly in perception tasks ~\cite{lu2023robust,wei2023asynchronyrobust,xiang2023v2xp,lang2019pointpillars,zhou_voxelnet_2018} crucial for safe navigation~\cite{li2023pretraining,xu2022pretram,huang2022survey,zhou2022comprehensive} and decision-making~\cite{wu2024cmp}. While progress in various perception tasks has been notable, the single-vehicle vision systems still suffer from occlusions and limited perception range. The bottlenecks mainly stem from the fact that each vehicle can only assess its surroundings from a single view point, leading to a partial understanding of the environment. To address these issues, recent works~\cite{wang2020v2vnet,xu2022opv2v,yu2022dair,li2022v2x,xiang2023hm,lu2024extensible} have investigated Vehicle-to-Everything (V2X) Cooperative Perception, where each connected agent within the communication range can share the sensing information (\eg, raw point clouds, detection results and intermediate neural features) with each other and leverage neighboring agent's sensing information for visual reasoning to enlarge the perception range and see through occlusions. 

Despite the great potential, there is no open V2X dataset from real-world scenarios for researchers to work with. The majority of available V2X datasets such as OPV2V~\cite{xu2022opv2v}, V2X-Sim~\cite{li2022v2x} and V2X-Set~\cite{xu2022v2x} are generated through simulators~\cite{dosovitskiy2017carla} with simulated traffic dynamics and sensor renderings. This approach introduces a significant simulation-to-reality gap, thereby impeding the effective evaluation and deployment of V2X algorithms in the real world~\cite{xu2023v2v4real}. Recently, DAIR-V2X~\cite{yu2022dair} and V2V4Real~\cite{xu2023v2v4real} present the pioneering real-world datasets for Vehicle-to-Infrastructure (V2I) and Vehicle-to-Vehicle (V2V) collaborations. However, these datasets are limited by a single collaboration mode (\ie either V2V or V2I), involving at most two agents within the same spatial vicinity. Nevertheless, in real world, V2X collaborations can dynamically adapt to multifaceted interactions inherent in real-world traffic scenarios, including V2V, V2I, I2I and thus truly V2X. Developing algorithms based on single collaboration mode will greatly constrain the scope of V2X collaborations and hinder deploying the algorithms in complex V2X scenarios. In addition, existing datasets are vehicle-centric by assuming the vehicle as the ego agent, ignoring the pivotal role of infrastructure; however, the infrastructure perception is vital for both autonomous driving and intelligent transportation systems.  On the one hand, it can augment the perception systems of self-driving cars by sharing sensing information with nearby agents. On the other hand, infrastructure perception can provide a reliable basis for various intelligent transportation tasks such as traffic management, traffic monitoring and signal control, which will greatly benefit from the V2I and I2I collaborations.  Moreover, DAIR-V2X~\cite{yu2022dair} and V2V4Real~\cite{xu2023v2v4real} are mostly collected in suburban and rural areas with an average of 11-12 objects per scene. This relatively low traffic density limits their effectiveness in testing V2X cooperative perception in more challenging scenarios and restricts the scalability of V2X collaborations. 
\begin{figure}[!t]
    \centering
     \includegraphics[width=0.97\columnwidth]{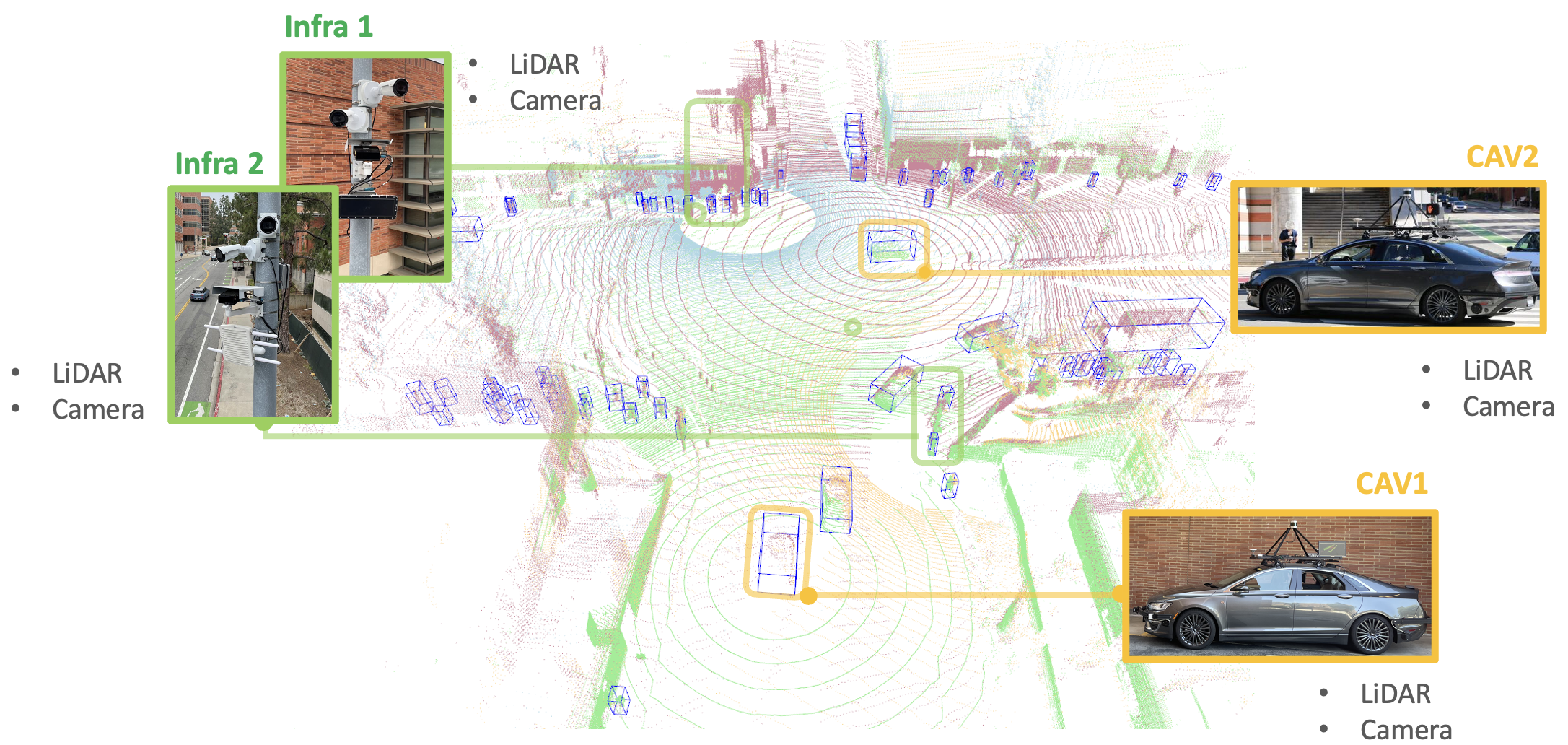}
    \caption{
    Demonstration of smart intersection and overall data acquisition systems. Smart infrastructure (Infra) units and connected and automated vehicles (CAVs) are depicted in their combined LiDAR point clouds. }
    \label{fig:pipeline}
\end{figure}

To this end, we introduce the V2X-Real, a large-scale multi-modal multi-view dataset tailored for V2X research, including 33K LiDAR frames, 171K camera images and over 1.2M 3D bounding box annotations. The proposed dataset is collected via two smart infrastructure and two autonomous vehicles equipped with full sensor suits (Fig.~\ref{fig:pipeline}). To ensure the data diversity, we gather the data in two types of scenarios \ie, V2X smart intersection and V2V corridors. With over 75 hours of collected driving logs, we carefully curate and select 68 representative scenarios to form the final dataset. Unlike previous datasets, V2X-Real is primarily collected in urban environments, featuring an average of 36 objects per scene. Such environments offer a high density of traffic and a significant presence of vulnerable road users, providing rich challenging scenarios for V2X cooperative perception research. To innovate and support a wide range of research interests in V2X cooperative perception, the dataset is divided into four specialized sub-datasets based on the type of ego agent and their collaboration modes: V2X-Real-VC for {V}ehicle-{C}entric, V2X-Real-IC for {I}nfrastructure-{C}entric, V2X-Real-V2V for {V}ehicle-to-{V}ehicle, and V2X-Real-I2I for {I}nfrastructure-to-{I}nfrastructure cooperative perception. 

The key contributions can be summarized as follows:
\begin{itemize}
    \item We build V2X-Real, the first open large-scale real-world dataset designed for V2X cooperative perception. According to ego agent's type and its collaboration mode, we derive 4 sub-datasets tailored for vehicle-centric, infrastructure-centric, V2V and I2I cooperative perception. 
    \item We provide over 1.2M annotated bounding boxes of 10 object categories with 33K LiDAR frames and 171K multi-view camera data.
    \item We conduct comprehensive benchmarks for multi-class multi-agent V2X cooperative perception for V2X-Real-VC, V2X-Real-IC, V2X-Real-V2V, and V2X-Real-I2I.  
\end{itemize}

\begin{table}[]
    \caption{Comparisons of our dataset and existing autonomous driving and V2X dataset. V2V: Vehicle-to-Vehicle. V2I: Vehicle-to-Infrastructure. I2I: Infrastructure-to-Infrastructure. VC: Vehicle-Centric. IC: Infrastructure-Centric. }
    \label{tab:dataset_comp}
    \centering  
    \begin{tabular}{l|c|c|c|c|c|c|c|c|c|c|c}
    \toprule
         \multirow{2}{*}{Dataset}&\multirow{2}{*}{Year}&\multirow{2}{*}{Type}&\multirow{2}{*}{V2V}&\multirow{2}{*}{V2I}&\multirow{2}{*}{I2I}&Image& Agent&RGB&LiDAR&3D&\multirow{2}{*}{Types}\\
         &&&&&&($360^\circ$)&Number&images&frames&boxes&\\
         \toprule
         KITTI~\cite{geiger2012we}   & 2012 & Real &          &            & & & 1 & 15k &15k&200k&8 \\ 
         nuScenes~\cite{caesar2020nuscenes}& 2019 & Real &          &            & & \checkmark& 1 & 1.4M&400k&1.4M&23\\
         Waymo Open~\cite{sun2020scalability}& 2019 & Real &        &            & & \checkmark& 1 & 1M  &200k& 12M&4\\ 
         \midrule
         OPV2V~\cite{xu2022opv2v}   & 2022 & Sim &\checkmark &            & & \checkmark& 2.89 & 44k &11k&230k&1\\
         V2X-Sim~\cite{li2022v2x} & 2022 & Sim &\checkmark & \checkmark & & \checkmark& 10 & 0   &10k&26.6k&1\\
         V2XSet~\cite{xu2022v2x}  & 2022 & Sim &\checkmark & \checkmark & & \checkmark& 2-7 & 44k &11k&230k&1\\
        \midrule
         DAIR-V2X~\cite{yu2022dair} &2022&Real&&\checkmark&&&2&39K&39K&464K&10\\
         V2V4Real~\cite{xu2023v2v4real} &2023&Real&\checkmark&&&&2&40K&20K&240K&5\\
         \midrule
         V2X-Real (ours) &2024&Real&\checkmark&\checkmark&\checkmark&\checkmark&\textbf{4}&\textbf{171K}&33K&\textbf{1.2M}&\textbf{10}\\
         -- V2X-Real-VC&2024&Real&\checkmark&\checkmark&&\checkmark&{4}&{145K}&30K&{1.1M}&{10}\\
         -- V2X-Real-IC&2024&Real&&\checkmark&\checkmark&\checkmark&{4}&{145K}&30K&{1.1M} &{10}\\
         -- V2X-Real-V2V&2024&Real&\checkmark&&&\checkmark&{2}&{140K}&17K&{719K}&{10}\\
         -- V2X-Real-I2I&2024&Real&&&\checkmark&&{2}&{31K}&15K&{470K}&{10}\\
         \bottomrule
    \end{tabular}
\end{table}

\section{Related Work}

\subsection{Self-driving datasets}
Over the past few decades, a number of autonomous driving datasets have been released, greatly promoting the research and development of self-driving technologies. Among these datasets, KITTI is a pioneering autonomous driving dataset featured with full sensor suits~\cite{geiger2012we}. Following KITTI, the NuScenes~\cite{caesar2020nuscenes} and Waymo Open datasets~\cite{sun2020scalability} have expanded the scale of data significantly, offering both LiDAR point clouds and multi-view cameras that cover a 360-degree field of view, thereby enhancing the depth and breadth of research possibilities in the field. However, these datasets are all collected by a single vehicle. OPV2V introduces the first V2V cooperative perception dataset, created using CARLA simulator~\cite{dosovitskiy2017carla} and OpenCDA~\cite{xu2021opencda,xu2023opencda,zheng2023opencda}. V2XSet and V2X-Sim~\cite{li2022v2x} further extended to V2X scenarios in the CARLA simulator. In contrast to these simulated datasets, DAIR-V2X~\cite{yu2022dair} is the first real-world dataset for V2I cooperative perception while V2V4Real~\cite{xu2023v2v4real} presents the first real-world V2V dataset collected by two connected vehicles.

Nevertheless, existing V2X datasets all focus on a single collaboration mode. Furthermore, the benchmarks and evaluations within these datasets are vehicle-centric, treating infrastructure merely as an auxiliary collaborator. Infrastructure-centric collaborations such as V2I and I2I are rarely discussed, highlighting a gap in the comprehensive evaluation of V2X interactions. A concurrent work, RCooper~\cite{hao2024rcooper}, introduces an I2I dataset for roadside cooperative perception. Compared with these datasets, our dataset features with all three V2X collaboration modes \ie V2V, V2I, and I2I, and investigates and benchmarks both Vehicle-Centric and Infrastructure Centric V2X cooperative perception. Moreover, it is uniquely collected in congested urban environments with dense traffic and pedestrian flow, resulting in a record number of annotated 3D boxes compared to existing datasets. Furthermore, to facilitate multi-view camera-based perception, we will also provide multi-view stereo cameras covering 360 field-of-view for the vehicle agent and high-resolution network camera data for the infrastructure. Additionally, existing V2X datasets only benchmark the detection performance based on a single vehicle category, ignoring the perception performance for vulnerable road users. In contrast, we provide benchmarks for all the annotation types and will release our codebase for multi-class cooperative perception to facilitate V2X research on vulnerable road users, which is an important yet under-explored research area.

\begin{table}[]
    \centering
     \caption{Key Sensor Specifications in V2X-Real Dataset. Infra means infrastructure.}
    \begin{tabular}{c|c|c|p{5.3cm}}
    \toprule
         \textbf{Agent}&\textbf{Sensor}&\textbf{Sensor Model} & \textbf{Details}  \\
         \toprule
         \multirow{3}{*}{Infra}&LiDAR& Ouster OS1-128/64 ($\times$1) & 128/64 beams, 10Hz capture frequency, $360^\circ$ horizontal FOV, $-16.6^\circ$ to $+16.6^\circ$(64), $-22.5^\circ$ to $+22.5^\circ$(128) vertical FOV, $\leq$ 40m range\\
         &Camera&Axis-P14555 ($\times$2) & RGB, 30Hz capture frequency, 1920 $\times$ 1080 resolution, $76^\circ$ horizontal FOV, $48^\circ$ vertical FOV, JPEG compressed\\ 
         &GPS& Garmin 18x LVC GPS ($\times$1) & 5Hz update rate\\
         \midrule
         \multirow{3}{*}{Vehicle}&LiDAR& RoboSense Ruby Plus ($\times$1) & 128 beams, 10Hz capture frequency, $360^\circ$ horizontal FOV, $-25^\circ$ to $+15^\circ$ vertical FOV, $\leq$ 200m range\\
         &Camera&ZED 2i ($\times$4) & Stereo RGBD, 10Hz capture frequency, 1920 $\times$ 1080 resolution, $120^\circ$ horizontal FOV, JPEG compressed\\
         &GPS/IMU & Gongji GPS System ($\times$1) & 1000Hz update rate, Double Precision\\
         \bottomrule
    \end{tabular}
    \label{tab:sensor_spec}  
\end{table}
\subsection{V2X Cooperative Perception}
Cooperative perception aims to enhance the perception capability of the V2X system by leveraging the shared information among connected agents. According to its fusion strategies, it can be broadly classified into three categories: 1) Early Fusion where the raw LiDAR point clouds are transmitted and the ego agent will aggregate the raw data to predict objects, 2) Late Fusion~\cite{rawashdeh2018collaborative} where the detection outputs are circulated and then fused into a consistent prediction, 3) Intermediate Fusion~\cite{wang2020v2vnet,xu2022v2x,xu2022opv2v,hu2022where2comm,li2021learning,su2023uncertainty,xu2023model}where the intermediate neural features are shared and fused.  Though early Fusion preserves the raw information and thus can potentially reach higher accuracy, the high bandwidth requirement of transmitting raw data makes it infeasible in real-world scenarios. On the other hand, Late Fusion requires minimal bandwidth but loses the context information during transmission, resulting in inferior performance with compound errors. To strike a good balance between bandwidth requirements and accuracy, Intermediate Fusion has been gaining increasing attention. AttFuse~\cite{xu2022opv2v} presents an attention-based fusion strategy to refine multi-agent's BEV features. V2VNet~\cite{wang2020v2vnet} employs a graph neural network architecture for joint perception and prediction. F-Cooper~\cite{chen2019f} presents a pooling mechanism to identify salient features. V2X-ViT~\cite{xu2022v2x} and CoBEVT~\cite{xu2022cobevt} employ a sparse vision transformer for V2X collaborations. HM-ViT~\cite{xiang2023hm} presents a 3D heterogeneous graph transformer for camera and LiDAR fusion.

\section{V2X-Real datasets}
To facilitate the V2X research, we present V2X-Real, the first large-scale real-world V2X dataset for cooperative perception. In this section, we first detail the data acquisition in Sec.~\ref{sec: data qcuisition}, then present the data annotation and processing pipeline in Sec.~\ref{sec:data_annotatiion}, and finally discuss the diverse data distribution and dataset statistics in Sec.~\ref{sec:dataset_analysis} 
\subsection{Data Acquisition}
\label{sec: data qcuisition}
\textbf{Sensor Specifications: }The dataset was gathered using two connected and automated vehicles and two smart infrastructure. Each of the vehicles is outfitted with a 128-beam LiDAR, four ZED 2i Stereo Cameras, and a GPS/IMU integration system. The four cameras are strategically positioned on the front, rear, left and right sides of the vehicle to achieve a full 360-degree field-of-view (FoV). Similarly, each infrastructure unit is equipped with a 128(64)-beam LiDAR, two Axis high-resolution network cameras and a GPS to ensure precise time synchronization among the distributed devices. Detailed specifications of these sensors are provided in Tab.~\ref{tab:sensor_spec}.

\noindent \textbf{Data Collection: }
To ensure the diversity of our V2X data, We collected our data in two types of scenarios --  V2X intersection and V2V corridor. In the intersection scenarios, two smart infrastructure units and two autonomous vehicles are operated simultaneously. Our autonomous vehicles will follow predefined routes, covering different V2X interactions. Through a combination of basic maneuvers of each autonomous vehicle at the intersection -- such as left turn, right turn, waiting for a traffic signal, and going straight -- we designed a total of 50 distinct V2X interactions in our dataset including car following, two cars driving towards each other, two cars making left (right) turn together \etal  As the V2X data is primarily collected in a single smart intersection and its surrounding streets for diverse V2X interactions, to enlarge the diversity of our dataset, we further collected separate V2V scenarios in urban corridor. In these V2V data, we designed 10 types of V2V interactions, including car following, lane change, and overtaking between two autonomous vehicles \etal. 

Our data collection efforts span 3 months in diverse illumination conditions, with a total of over 75 hours of collected driving logs. We carefully curate the data and select 68 representative scenarios for annotation with a total of 33K LiDAR frames and 171K Camera images, providing a rich resource for V2X research. 
\begin{figure*}[!t]
\centering
    \begin{subfigure}[c]{0.49\linewidth}
        \centering{\includegraphics[width=1\linewidth]{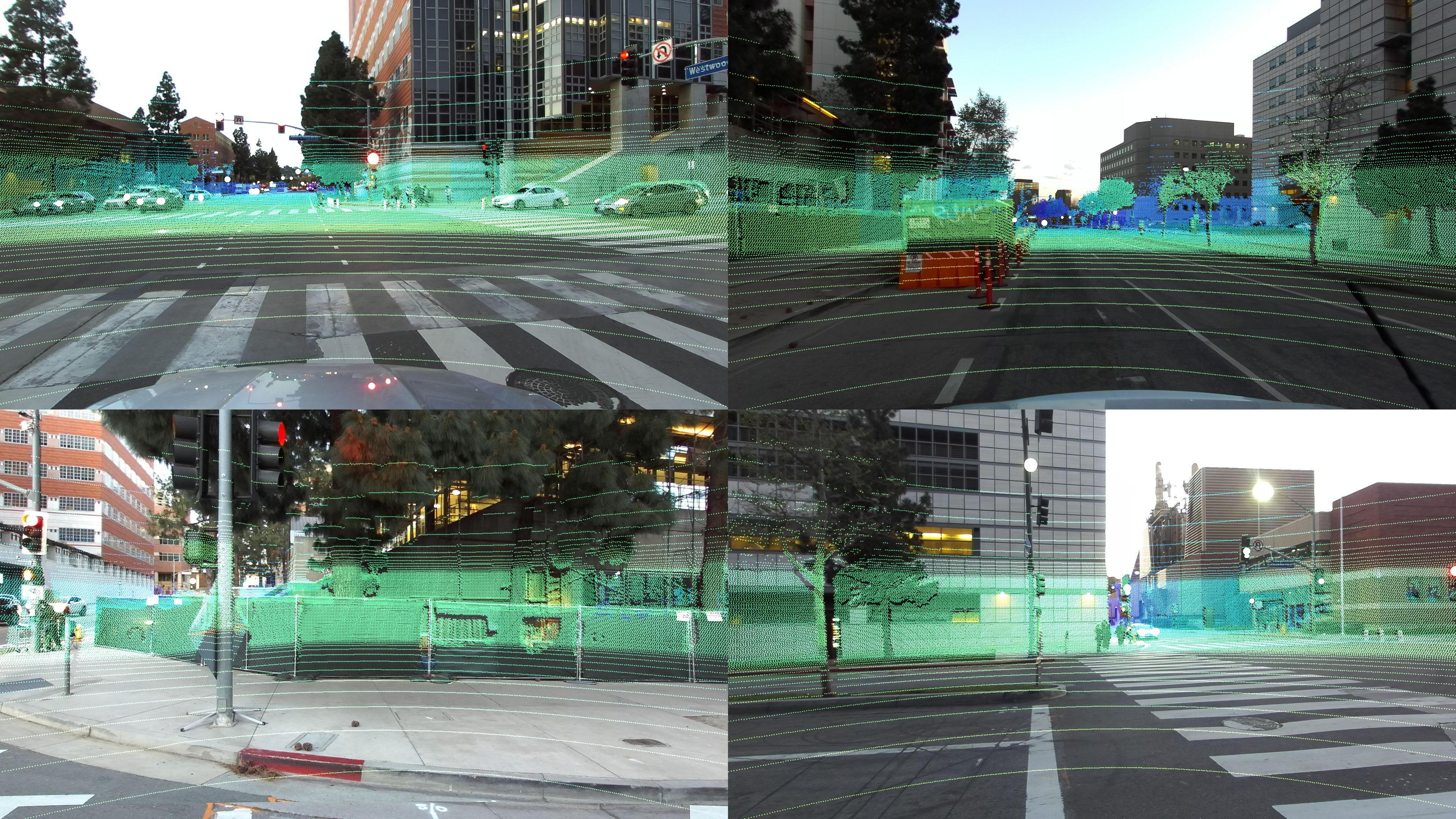}}
        \caption{Vehicle calibration results}
        \label{fig:calib-a}
    \end{subfigure}
    \begin{subfigure}[c]{0.49\linewidth}
        \centering{\includegraphics[width=1\linewidth]{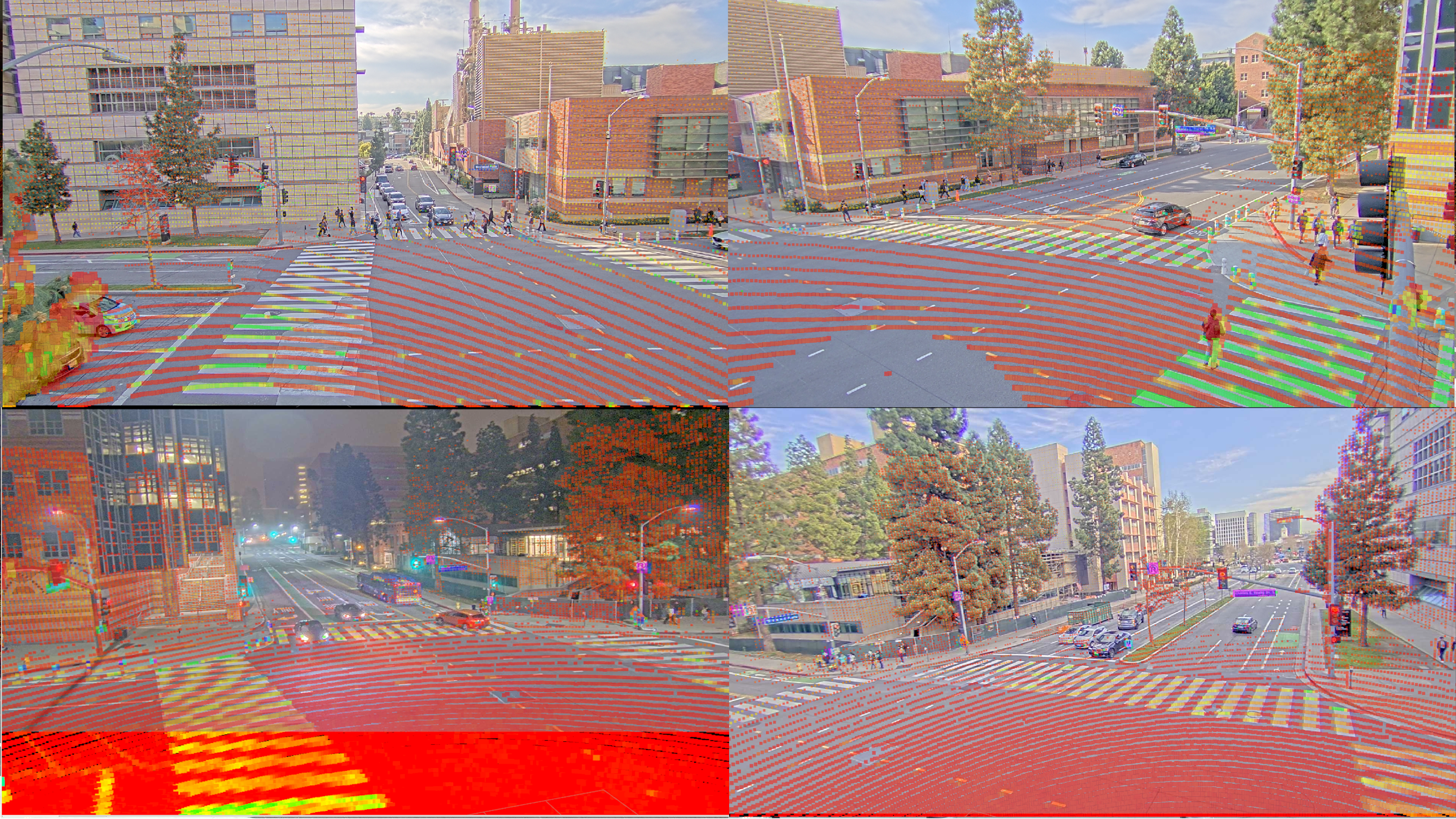}}
        \caption{Infrastructure calibration results}
        \label{fig:calib-b}
    \end{subfigure}
    \caption{Visualization of calibration results for (a) vehicle and (b) infrastructure (better viewed in color and zoomed in). The LiDAR points are projected onto the camera plane using the camera intrinsics and camera-LiDAR extrinsics.}
    \label{fig:calib}
\end{figure*}

\begin{figure}
    \centering
\includegraphics[width=0.8\columnwidth]{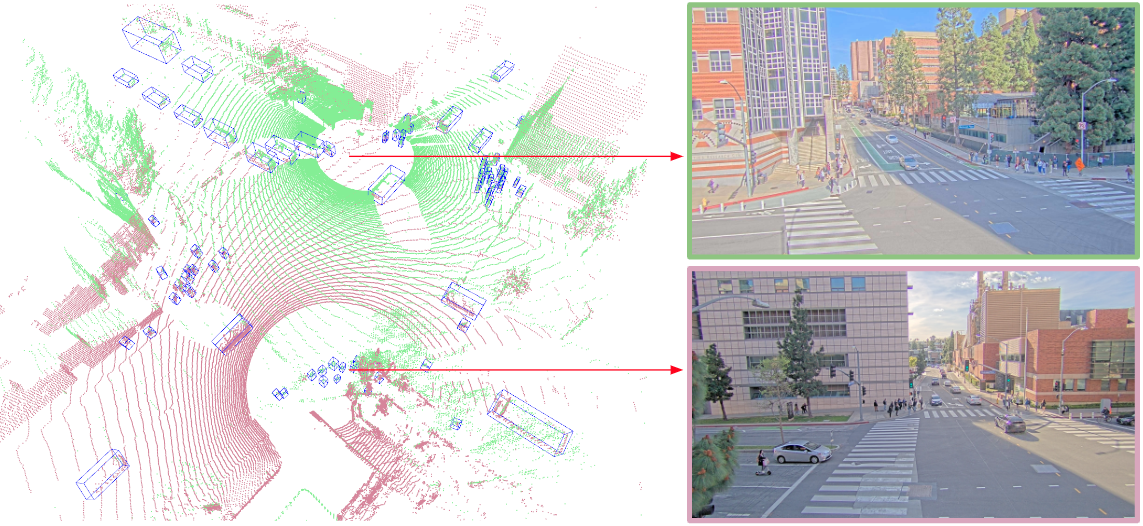}
    \caption{Demonstration of combined LiDAR annotations for infrastructure. LiDAR points from different agents (\textcolor{green}{green} and \textcolor{pink}{pink}) are aggregated for the same object, reducing half of the infrastructure annotation workload and enhancing annotators' ability to identify objects accurately and efficiently. }
    \label{fig:annotation}
\end{figure}
\noindent \textbf{Coordinate system: }There are three types of coordinate systems in our dataset, namely the LiDAR coordinate, the camera coordinate, and the map coordinate. Each agent maintains a separate LiDAR and camera coordinate frame. We built a dense LiDAR point cloud map~\cite{xia2023automated} covering all the driving routes and leverage map matching combined with GPS/IMU positional inputs~\cite{gao2023gnss} to identify the transformation between each agent's LiDAR coordinate frame and map coordinate. For 3D-2D associations, intrinsic and extrinsic calibrations between the camera and LiDAR are conducted. The visualization of the calibration result is shown in Fig.~\ref{fig:calib}.

\noindent\textbf{Synchronization: }For the multi-agent datasets, it is crucial to synchronize the time clock of all agents. We synchronize all the computer's time clock with GPS time to ensure consistent timestamp across sensing messages and further lock the LiDAR phase with GPS signals to ensure the temporal synchrony of multi-agent sensing observation. As the ZED2i stereo camera and Axis network camera cannot be hardware-triggered with GPS signals for synchronization, we post-process the collected data 
to identify the closest camera message as per LiDAR frame to reach minimal time difference.  Messages with a large time gap are filtered to ensure suitable synchronization between multi-agent multi-modal data.

\subsection{Data Annotation and Processing}
\label{sec:data_annotatiion}
We leverage the open-sourced labeling tool called SUSTechPOINTS~\cite{li2020sustech} to annotate 3D bounding boxes. Expert annotators are employed to make high-quality annotations with a stringent review and revision process in place. On average, each scenario goes through 5-6 rounds of review and revision to guarantee accuracy and quality. In our dataset,  there are in total 10 categories of annotated objects including pedestrian, scooter, motorcycle, bicycle, car, truck, van, box truck, bus and barrier on the road. For each object, we annotate its 9-dimensional cuboid containing x, y, z for the cuboid center, width, length, height for the cuboid extent dimension, and roll, yaw, pitch for the cuboid orientation.

\begin{figure}
    \begin{subfigure}{0.328\linewidth}
    \includegraphics[height=1.6in]{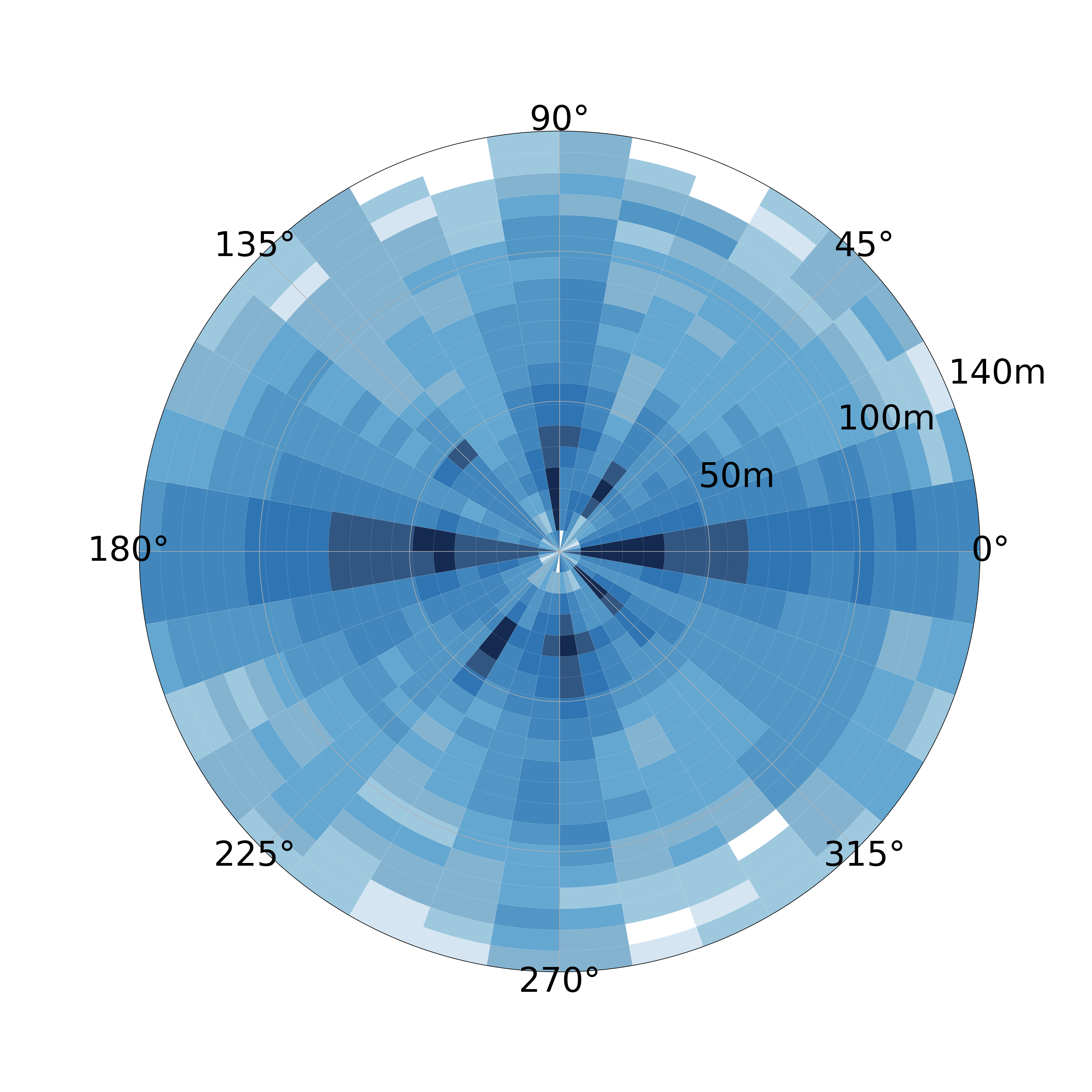}
    \caption{Car}
    \label{fig:box_polar-a}
  \end{subfigure}
  \hfill
  \begin{subfigure}{0.328\linewidth}
    \includegraphics[height=1.6in]{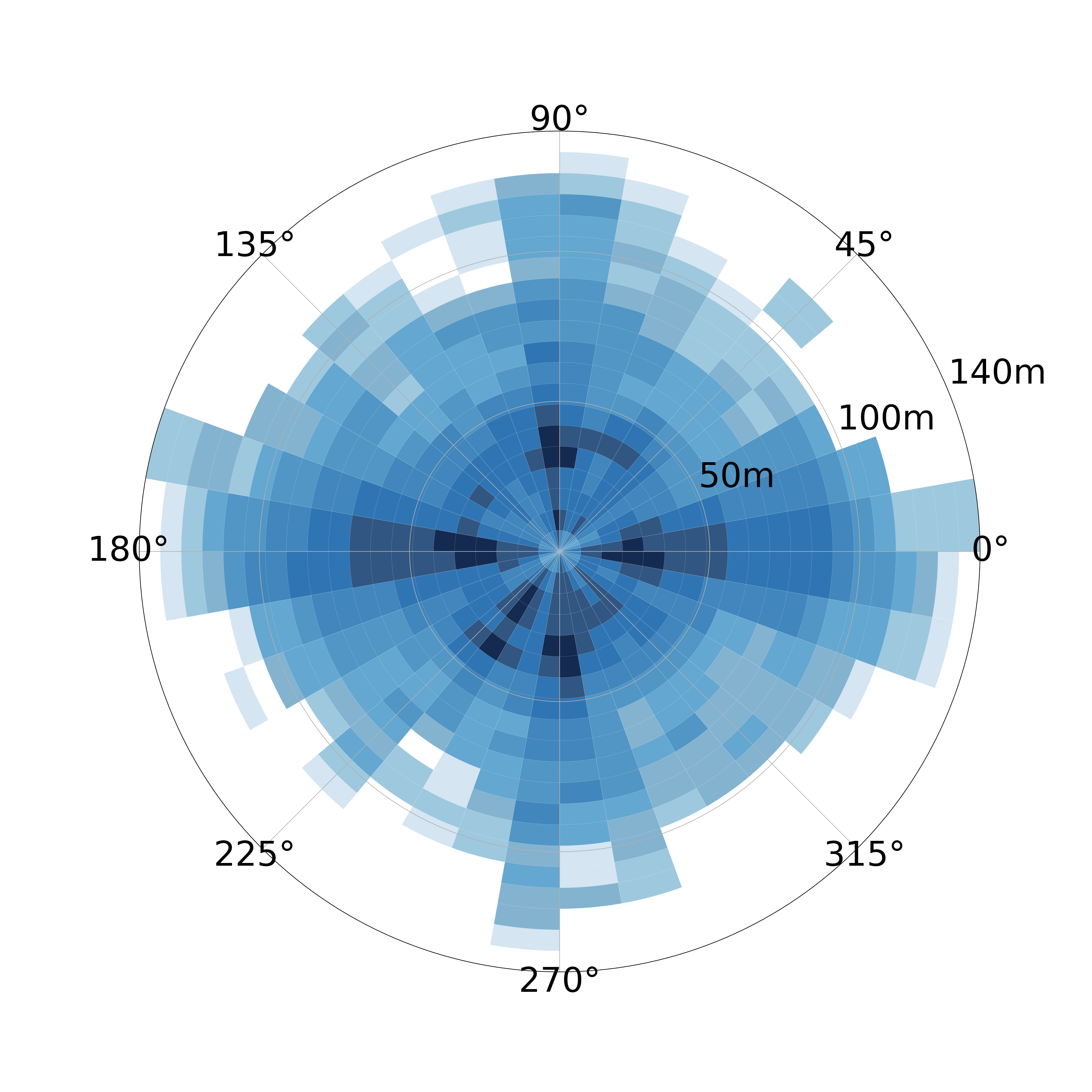}
    \caption{Pedestrian}
    \label{fig:box_polar-b}
  \end{subfigure}
  \hfill
  \begin{subfigure}{0.328\linewidth}
    \includegraphics[height=1.6in]{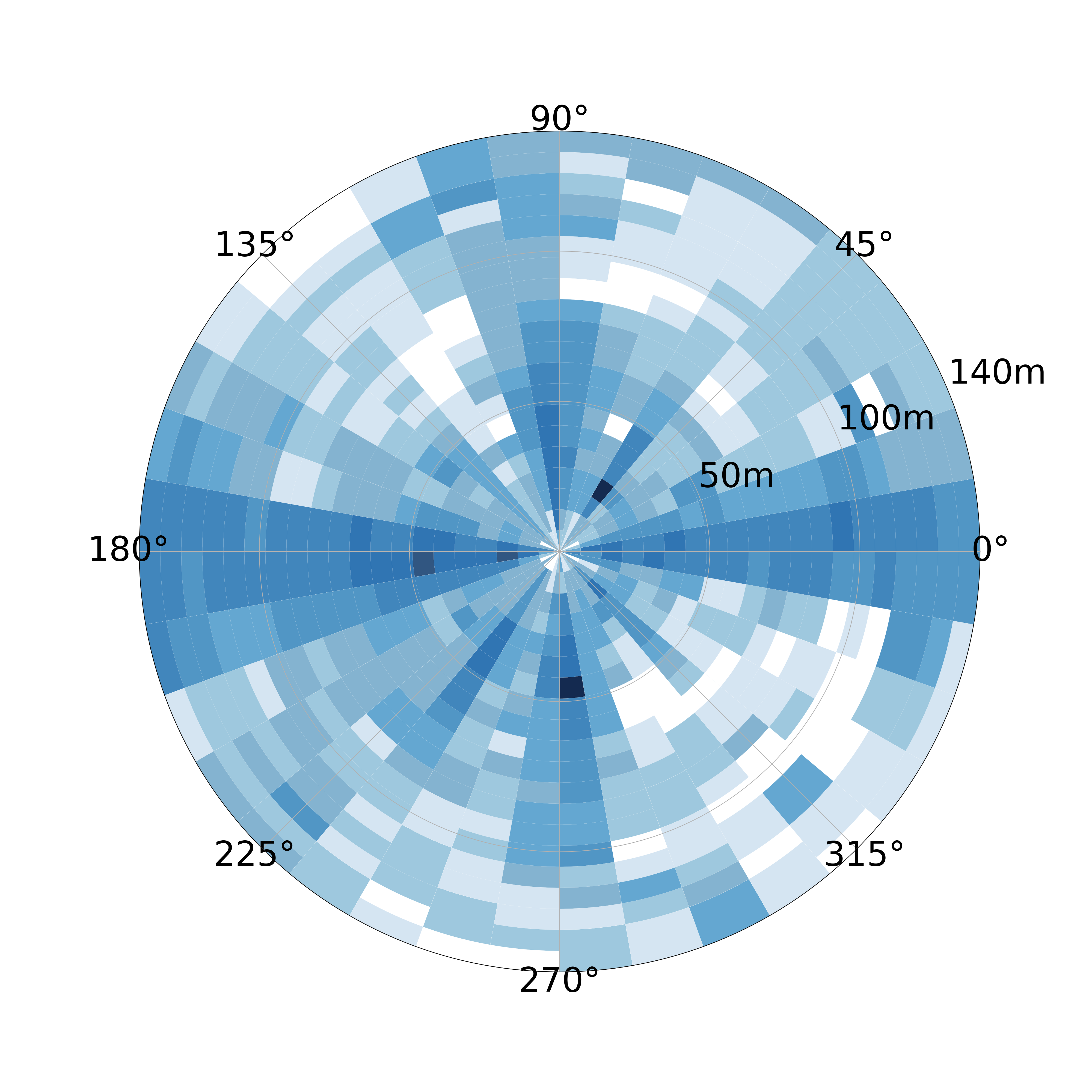}
    \caption{Truck}
    \label{fig:box_polar-c}
  \end{subfigure}
  \caption{Polar density map of annotated bounding boxes for (a) Car, (b) Pedestrian, and (c) Truck. The polar and radial axes represent the distance and angle of the bounding boxes respectively in the ego coordinate frame. The color indicates the number of boxes in the bin and the darker color corresponds to a larger density.  }
  \label{fig:box_polar}
\end{figure}

Additionally, we have implemented several strategies to enhance the efficiency and accuracy of the annotation process. We combine two infrastructure's LiDAR point clouds to form a holistic view for annotation (Fig.~\ref{fig:annotation}). The combined LiDAR point clouds can efficiently reduce the annotation workload and provide dense sensing measurements for objects of interest, greatly increasing the annotation efficiency and accuracy. After annotating the 3D boxes in combined LiDAR point clouds,  we will project the boxes into each infrastructure's coordinate frame and for each box, we will count the number of LiDAR points within the box. If the number is greater than a threshold (\ie, 5 in our dataset), we will add the box to this infrastructure's annotation result, otherwise, the box is ignored for this agent due to its sparse sensing measurements. Afterwords, we manually check the result and refine the annotations for each of the agents. For vehicles, their poses vary from scene to scene with different levels of localization error, thus separate annotations for each of the vehicles are conducted to ensure accurate annotations. Additionally, for the static objects (\eg, vehicles that are parked and those awaiting the change of a traffic signal), we provide annotators with a script to automatically interpolate their poses in one frame to a sequence of frames. This greatly increased the annotation accuracy and efficiency for static objects.

\begin{figure}
    \centering
     \includegraphics[width=0.4\columnwidth]{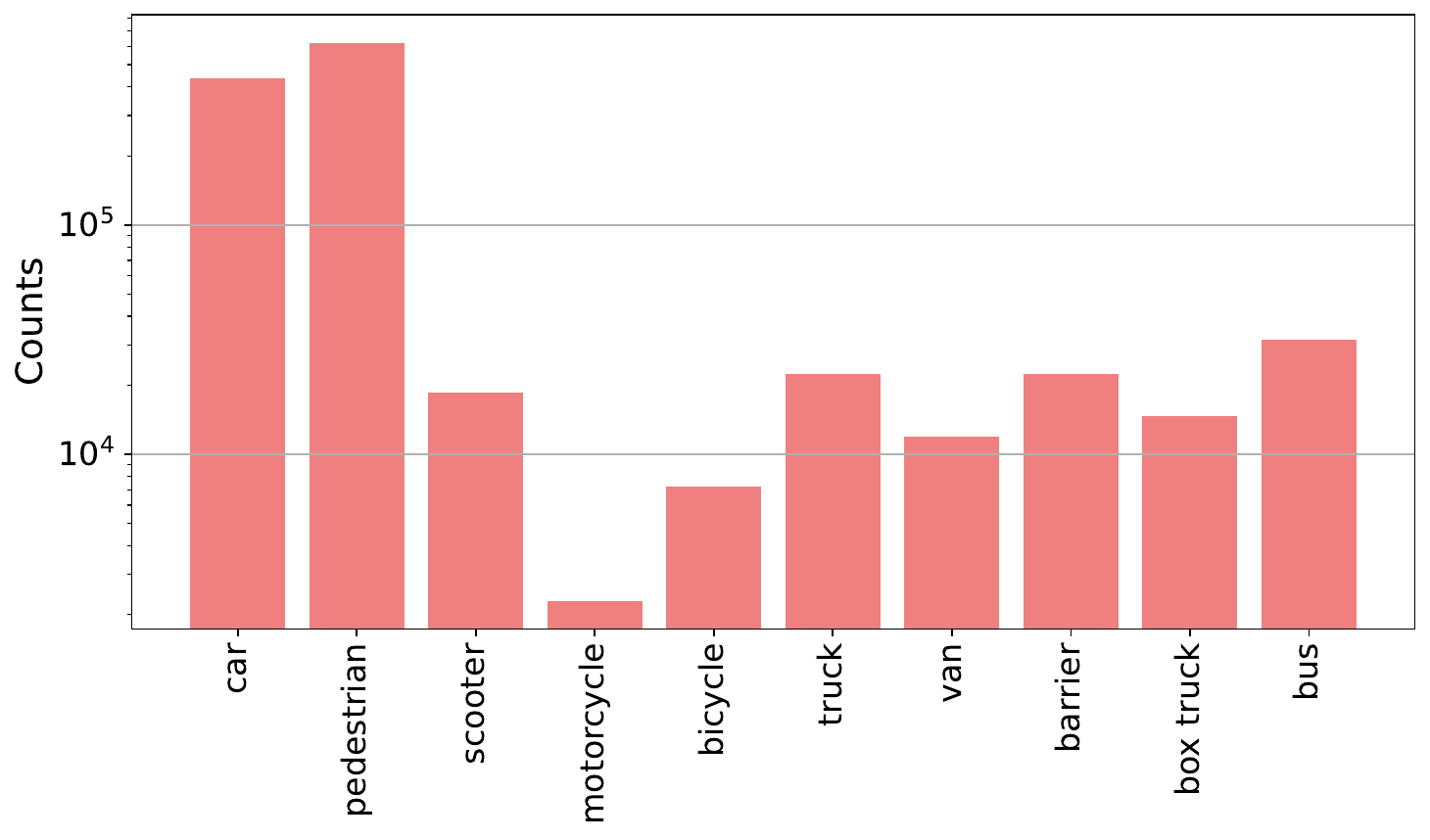}
    \caption{Number of annotations per category in log scale. }
    \label{fig:number_of_category}
\end{figure}

As the objects are annotated separately in different agents' coordinate frames, identical objects may be assigned with different IDs. To ensure consistent object ID, we project all the annotated boxes from different agents to the same map coordinate and calculate their IoU overlap. If their overlap is greater than a certain threshold, they are considered the same object and thus a unique ID is assigned.

\begin{figure}
    \begin{subfigure}{0.33\linewidth}
    \includegraphics[height=1.22in]{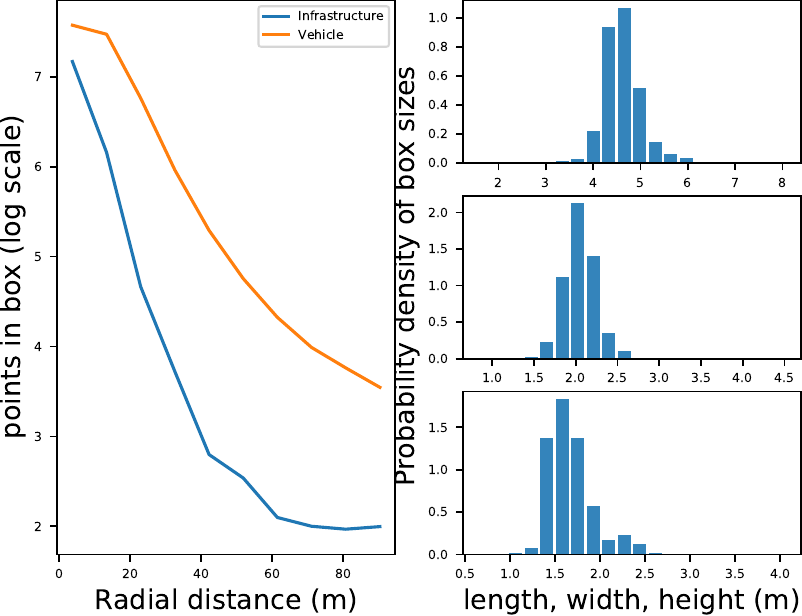}
    \caption{Car}
    \label{fig:box_size-a}
  \end{subfigure}\hfill
  \begin{subfigure}{0.33\linewidth}
    \includegraphics[height=1.22in]{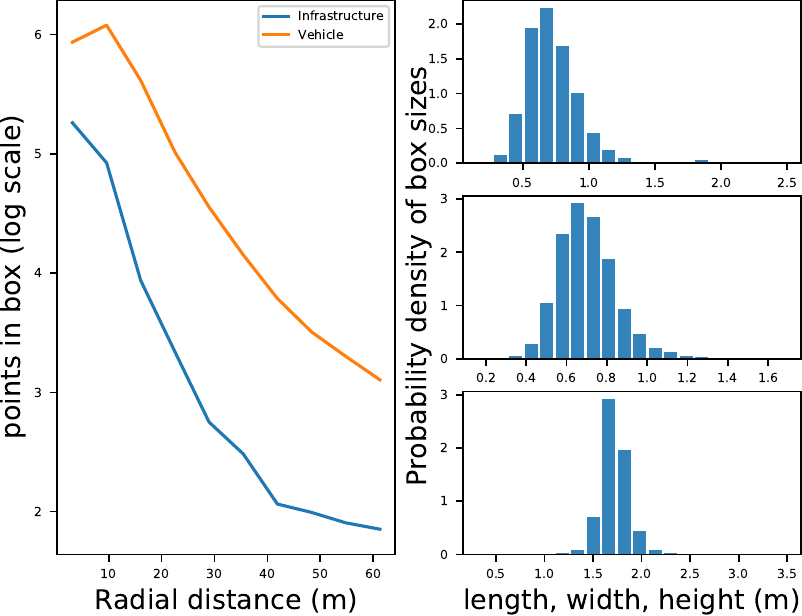}
    \caption{Pedestrian}
    \label{fig:box_size-b}
  \end{subfigure}\hfill
  \begin{subfigure}{0.33\linewidth}
    \includegraphics[height=1.22in]{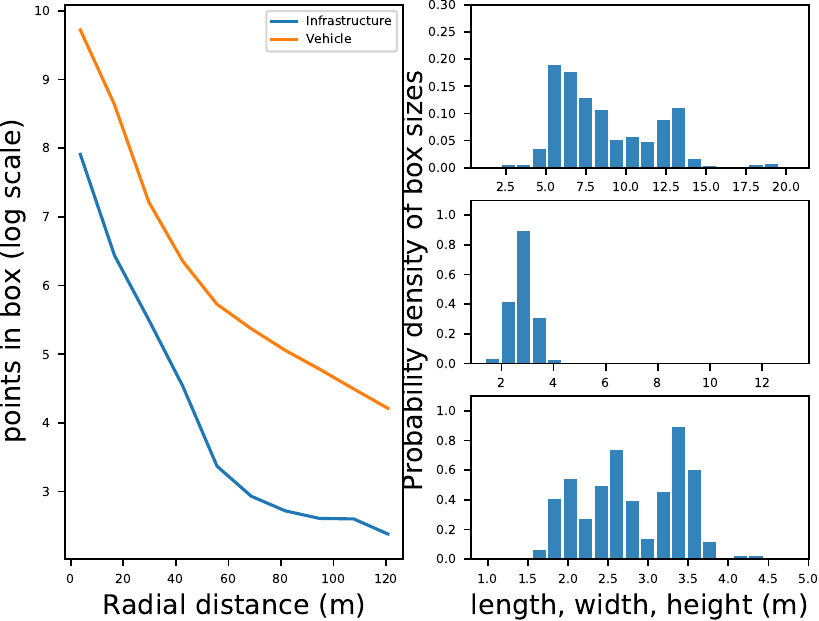}
    \caption{Truck}
    \label{fig:box_size-c}
  \end{subfigure}
  \caption{Bounding box LiDAR point density distribution and box size distribution for (a) Car, (b) Pedestrian, and (c) Truck. In each category's figure, the left figure depicts the number of lidar points within bounding boxes as per the radial distance from the data collection vehicle while the right figure demonstrates the bounding box size distribution. }
  \label{fig:box_size}
\end{figure}

\subsection{Dataset Analysis}
\label{sec:dataset_analysis}
Our dataset has an average of 36 objects per scene with a maximum of 115 and a minimum of 9 objects, ensuring a high density of traffic in challenging urban scenarios. The number of annotations for each of the 10 categories is shown in Fig.~\ref{fig:number_of_category}. The dominant objects in the dataset are pedestrians and cars. Compared with the existing V2X dataset (Tab.~\ref{tab:dataset_comp}), V2X-Real is featured with a wide variety and quantity of vulnerable road users including pedestrian, scooter, motorcycle, and bicycle. In our benchmarks, we group the 10 categories, based on their box sizes, into three super-classes \ie pedestrian, car, and truck. In Fig.~\ref{fig:box_polar}, we reveal the polar density map of annotated bounding boxes for these three classes, where each bin in the polar map represents a spatial area with a certain distance and angle range in the ego agent's coordinate frame.  From the figure, we can observe a diverse distribution of annotations and each category is denser in the front, left, right, and rare sides of the ego vehicle. We argue this is due to the fact that most of our data is collected adjacent to the intersection and the surrounding vehicles tend to have these four relative poses with respect to the data collection vehicle. Compared with car and truck categories, pedestrians have fewer annotations above 100 meters due to the sparse LiDAR measurements of the small objects in distant areas. Fig.~\ref{fig:box_size} depicts the point clouds density distribution within bounding boxes and the bounding box size distribution. From the figure, we can see the diverse bounding box size distribution and each category will have a size distribution and could serve as a prior for object detection tasks. Additionally, the number of points within the box will decrease for longer radial distances while collaboratively viewing the objects can potentially increase the sensing observation and thus provide better visual cues. 

\section{Tasks}
In this section, we first present the V2X cooperative 3D object detection task with derived datasets as per the collaboration mode and ego perspective in Sec.~\ref{sec:detection}, and then we introduce the evaluation metrics in Sec.~\ref{sec:metrics}. Afterwords, in Sec.~\ref{sec:benchmark}, we introduce various benchmark methods and fusion strategies supported in our dataset. 
\subsection{V2X Cooperative 3D object detection}
\label{sec:detection}
\textbf{Scope: }The V2X Cooperative  3D object detection aims to leverage multi-agent's LiDAR point clouds to collaboratively perform 3D object detection. In our dataset, there are four collaborators including two infrastructure and two autonomous vehicles. Based on the V2X collaboration mode and ego agent type, we present four types of datasets: V2X-Real-VC, V2X-Real-IC, V2X-Real-V2V, and V2X-Real-I2I.

\noindent\textbf{V2X-Real-VC: }In V2X-Real-VC, the ego agent is fixed as the autonomous vehicle while the collaborators include both the infrastructure and vehicle. This setup is specifically designed for research of vehicle-centric V2X collaborations. 

\noindent\textbf{V2X-Real-IC: }The infrastructure is chosen as the ego agent and the neighboring vehicles and infrastructure can collaborate with the ego infrastructure via sharing sensing observations. The final evaluation is conducted in the ego infrastructure side. This dataset is tailored for infrastructure-centric cooperative perception study. 

\noindent\textbf{V2X-Real-V2V: }Only Vehicle-to-Vehicle collaboration is considered in this dataset. Additional scenarios collected in V2V corridors are used to supplement intersection scenarios with more diverse data distribution.

\noindent\textbf{V2X-Real-I2I: }In this dataset, Infrastructure-to-Infrastructure collaborations are studied, which is critical for intelligent transportation systems.  

Contrary to existing works tailored for a single collaboration mode, the presented datasets have been meticulously crafted to cater to the diverse research interests of the broader V2X community. 

\subsection{Metrics}
\label{sec:metrics}
\noindent\textbf{Ground truth: }During the training stage, a random agent will be selected as the ego agent while during the inference, the ego is fixed as per dataset categories. For example, in the V2X-Real-VC, the ego is fixed as one of the data collection vehicles while in the V2X-Real-IC, the ego is infrastructure. When the bounding boxes are not consistent across agents, we choose the ego agent's annotation as the ground truth.

\noindent\textbf{Evaluation: }The evaluation is conducted in the range of -100m to 100m in x direction and -40m to 40m in y direction of the ego coordinate frame. We group the annotated 10 categories into three classes \ie car, pedestrian and truck as per their bounding box size distribution. We calculate the Average Precision(AP) for each class based on the specified Intersection-over-Union (IoU) threshold, and a final mean Average Precision (mAP) is evaluated based on each category's AP: 
\begin{equation}
    mAP=\frac{1}{C}\sum_{i=1}^{C}AP_i
\end{equation}
where $C$ is the class number. As our dataset contains a substantial number of pedestrians and large buses and the bounding box sizes can vary from less than 0.5 meters (pedestrian) to over 20 meters (bus), accurately predicting objects' poses and dimensions is particularly challenging. Thus similar as KITTI-360 dataset~\cite{liao2022kitti}, we adopt smaller IoU thresholds (IoU=0.3, 0.5) for the AP calculation to faithfully reflect model's perception capability. 
\begin{table}[]
    \centering
    \caption{Benchmark results of SOTA cooperative perception methods for V2X-Real-VC, V2X-Real-IC, V2X-Real-V2V, and V2X-Real-I2I. All numbers in the table are represented in percentage format. For detection accuracy, Average Precision (AP) and mean Average Precision (mAP) for cars, pedestrians, and trucks are reported under IoU thresholds of 0.3 and 0.5. }
    \begin{tabular}{c|c|cccccc|cc}
    \toprule
       \multirow{2}{*}{Dataset} &\multirow{2}{*}{Models}& \multicolumn{2}{c}{AP$_{car}$@IoU} & \multicolumn{2}{c}{AP$_{ped.}$@IoU} &  \multicolumn{2}{c|}{AP$_{truck}$@IoU}& \multicolumn{2}{c}{mAP@IoU}  \\
        & & 0.3 & 0.5 & 0.3 & 0.5 & 0.3 & 0.5 & 0.3 & 0.5\\
        \midrule
        \multirow{6}{*}{V2X-Real-VC}&No Fusion& 38.7 & 35.9 & 25.5 & 13.1 & 20.2 & 14.5 & 28.2 & 21.2\\
        \cmidrule{2-10}
        &Late Fusion& 46.1 & 43.1 & 28.3 & 11.8 & 19.8 & 14.0 & 31.4 & 23.0  \\
        \cmidrule{2-10}
        &Early Fusion&51.1&47.6&31.6&16.0&32.5&23.6&38.4&29.1\\
        \cmidrule{2-10}
        &F-Cooper~\cite{chen2019f}&57.3&54.2&30.0&14.1&27.0&21.2&38.1&29.8 \\
        &AttFuse~\cite{xu2022opv2v}&62.6&59.4&32.2&15.5&32.6&26.6&42.5&33.8\\
        &V2X-ViT~\cite{xu2022v2x}&62.7&60.3&36.7&18.6&35.1&28.3&44.8&35.8 \\
        \midrule
        \multirow{6}{*}{V2X-Real-IC}&No Fusion&43.3&35.6&25.6&12.7&21.2&20.2&30.0&22.8 \\
        \cmidrule{2-10}
        &Late Fusion&65.2&60.4&33.4&16.0&30.8&24.5&43.1&33.6\\
        \cmidrule{2-10}
        &Early Fusion&64.2&59.8&32.3&15.0&33.3&27.4&43.3&34.1 \\
        \cmidrule{2-10}
        &F-Cooper~\cite{chen2019f}& 49.7&43.0&31.0&15.3&31.9&21.1&37.5&26.5\\
        &AttFuse~\cite{xu2022opv2v}&70.4&67.3&37.5&17.0&42.9&31.3&50.3&38.5 \\
        &V2X-ViT~\cite{xu2022v2x}&64.2&56.5&37.8&19.2&35.6&28.8&45.9&34.8\\
        \midrule
        \multirow{6}{*}{V2X-Real-V2V}&No Fusion&41.7&39.4&26.9&14.4&21.6&13.7&30.1&22.5 \\
        \cmidrule{2-10}
        &Late Fusion&47.4&44.4&29.2&14.9&18.7&9.1&31.8&22.8\\
        \cmidrule{2-10}
        &Early Fusion&54.0&49.8&31.9&17.1&28.6&18.6&38.1&28.5\\
        \cmidrule{2-10}
        &F-Cooper~\cite{chen2019f}&42.7&40.3&27.7&14.0&25.6&18.6&32.0&24.3 \\
        &AttFuse~\cite{xu2022opv2v}&58.6&55.3&30.1&15.4&28.9&21.7&39.2&30.8 \\
        &V2X-ViT~\cite{xu2022v2x}&{59.0}&{56.3}&{37.4}&{20.7}&{42.9}&{35.0}&{46.5}&{37.3}\\
        \midrule
        \multirow{6}{*}{V2X-Real-I2I}&No Fusion&48.6&40.0&30.9&15.8&23.5&22.4&34.3&26.1 \\
        \cmidrule{2-10}
        &Late Fusion&67.2&63.3&41.1&23.1&48.4&39.1&52.2&41.8\\
        \cmidrule{2-10}
        &Early Fusion&60.9&57.2&41.2&21.9&38.5&30.8&46.9&36.6 \\
        \cmidrule{2-10}
        &F-Cooper~\cite{chen2019f}&71.5&65.5&49.3&27.5&50.0&40.7&56.9&44.6\\
        &AttFuse~\cite{xu2022opv2v}&73.5&69.5&42.8&20.5&53.2&39.4&56.5&43.1 \\
        &V2X-ViT~\cite{xu2022v2x}&{77.3}&{68.8}&{54.4}&{30.5}&{56.2}&{51.5}&{62.7}&{50.3}\\
        \midrule
    \end{tabular}
    \label{tab:benchmark}
\end{table}
\subsection{Benchmark methods}
\label{sec:benchmark}
We provide benchmarks for all three fusion strategies in cooperative perception with SOTA fusion methods. Note that different from existing V2X 3D object detection benchmarks~\cite{yu2022dair, xu2023v2v4real} where only single vehicle categories are detected and evaluated, we provide comprehensive benchmarks for all the annotated classes \ie pedestrian, car, and truck, providing the first open cooperative perception framework for multi-class cooperative 3D object detection. Here we list the benchmarked fusion strategies and methods:  \\
\begin{itemize}
    \item \textbf{No Fusion: }Only ego agent's LiDAR will be used to produce the 3D bounding boxes. It serves as the baseline. 
    \item \textbf{Late Fusion: }Each agent first generates its own predictions and then shares the detection results with neighboring agents. The ego agent will receive these proposals and leverage non-maximum suppression (NMS) to produce consistent predictions.  
    \item \textbf{Early Fusion: }The raw LiDAR point clouds will be shared for agents within the communication range. After receiving the neighboring agent's LiDAR point clouds, the ego agent will transform LiDAR point clouds into the ego agent's coordinate frame and aggregate all the LiDAR point clouds to produce a holistic view for visual reasoning. 
    \item \textbf{Intermediate Fusion: }Each agent will first project its LiDAR to the ego agent's coordinate frame and leverage the LiDAR backbone to extract Birds' Eye View (BEV) features, which are then compressed and shared with the ego agent. The ego agent will leverage received multi-agent features to refine its own feature representations. In this work, we provide benchmarks for AttFuse~\cite{xu2022opv2v}, F-Cooper~\cite{chen2019f}, and V2X-ViT~\cite{xu2022v2x}. 
\end{itemize}

\section{Experiments}

\subsection{Implementation details }
We split the dataset into train/val/test with 23379, 2770, and 6850 frames respectively. For the detection models, we adopt PointPillar~\cite{lang2019pointpillars} as the LiDAR backbone. The cooperative perception framework is built upon OpenCOOD~\cite{xu2022opv2v}. To enable multi-class cooperative perception, we extend the original single-class head to multi-class heads following OpenPCDet's design~\cite{openpcdet2020}. All the models are trained with 80 epochs with a batch size of 2 and a learning rate of 0.001. Adam optimizer (decay weight $\lambda=10^{-4}$) and multi-step learning rate scheduler ($\gamma=0.1$, milestones=[10,50]) are employed for optimization.

\subsection{Benchmark results}
Tab.~\ref{tab:benchmark} summarizes the benchmark results of all six models for the four datasets. For all the datasets, cooperative methods outperform No Fusion baseline, demonstrating the great potential of V2X collaboration in enhancing the perception capabilities of individual agents. Among these cooperative methods, Intermediate Fusion methods typically yield more precise predictions compared to Late and Early Fusion strategies. Notably, across all three object categories, the models generally perform best in the car category, whereas the performance is comparatively lower for the pedestrian and truck categories. We argue this is due to the fact that truck type has a diverse dimension distribution (Fig.~\ref{fig:box_size-c}) and pedestrian bounding boxes usually only contain a few sparse LiDAR points, posing challenges for the learning. Generally, intermediate fusion ranks first among three fusion strategies, and late fusion methods outperform No Fusion baselines while early fusion shows superior performance than late fusion. 

\section{Conclusion}
In this work, we present the V2X-Real, the first large-scale real-world dataset for V2X cooperative perception. It encompasses 33K LiDAR frames, 171K RGB images, and an unprecedented 1.2 million annotated 3D bounding boxes, making it one of the largest V2X datasets of its kind with dense traffic flows. To facilitate a wide range of V2X collaboration research, we provide four sub-datasets including V2X-Real-VC, V2X-Real-IC, V2X-Real-V2V, and V2X-Real-I2I.  We also present a comprehensive multi-class benchmark framework for cooperative 3D object detection and from the best of our knowledge, this framework is also the first open multi-class cooperative 3D object detection framework in the literature, which enables future multi-class cooperative perception research like cooperative perception for vulnerable road users. We will release our data and benchmark codes and hope our open-source efforts can inspire more researchers to investigate this new yet important field.

\noindent\textbf{Future work and limitation: }The current dataset comprises only LiDAR and camera data, lacking support for radar, which is crucial for developing low-cost, scalable autonomous driving solutions. In the future, we plan to further extend our sensor suits to low-cost sensors such as radar and low-beam LiDAR for studying scalable low-cost V2X solutions. 

\section*{Acknowledgements}
Federal Highway Administration Center of Excellence on New Mobility and Automated Vehicles, National Science Foundation \# 2346267 POSE: Phase II: DriveX: An Open-Source Ecosystem for Automated Driving and Intelligent Transportation Research

\newpage
\appendix
\section*{Appendix}
In this appendix, we first present additional visualizations for the V2X-Real dataset (Sec.~\ref{sec:dataset_visualization}). Then, we detail the model configurations for LiDAR-based and Camera-based Cooperative Perception methods (Sec.~\ref{sec:model_details}). Afterwords, additional benchmark results, and qualitative visualizations are shown in Sec.~\ref{sec:exp}.

\begin{figure}[]
\centering
\footnotesize
\def\xwidth{0.88}
\begin{tabular}{c}
\includegraphics[width=0.75\linewidth]{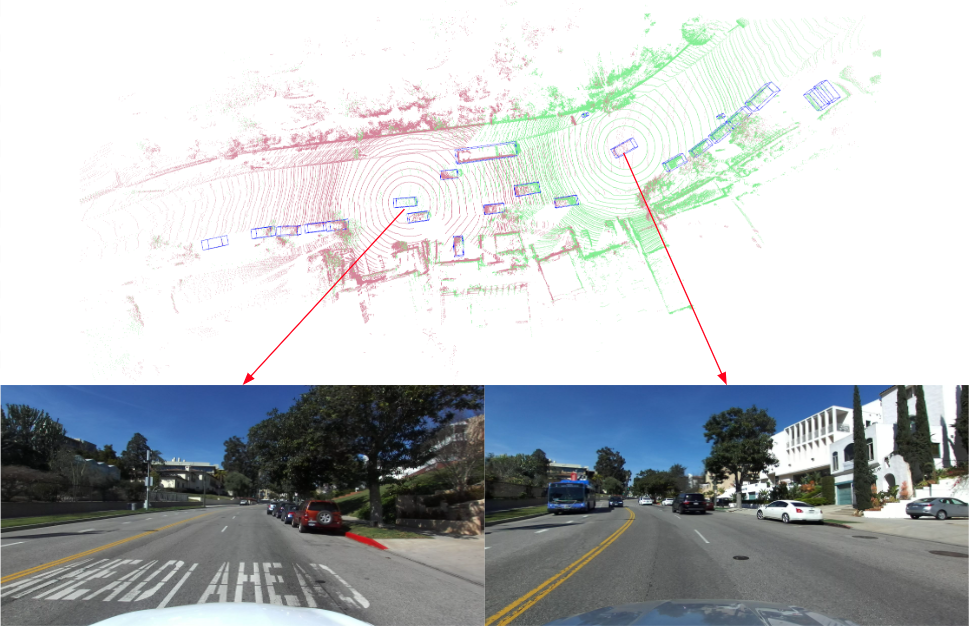} \\
(a) A V2V cityroad\\
\includegraphics[width=0.75\linewidth]{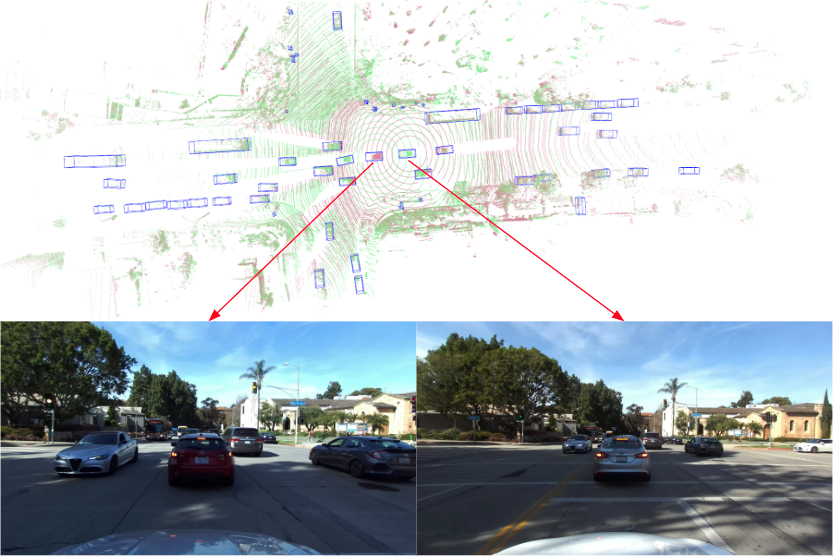}\\
(b) A V2V four-way intersection\\
\end{tabular}
\caption{Sample scenarios from V2V corridors. \textit{Top: }The aggregated LiDAR point clouds. \textit{Bottom: }The front cameras of two data collection vehicles. }
\label{fig:v2v_corridor}
\
\end{figure}

\section{Dataset Visualization}
\label{sec:dataset_visualization}
\begin{figure*}[]
\centering
    \begin{subfigure}[c]{0.49\linewidth}
        \centering{\includegraphics[width=1\linewidth]{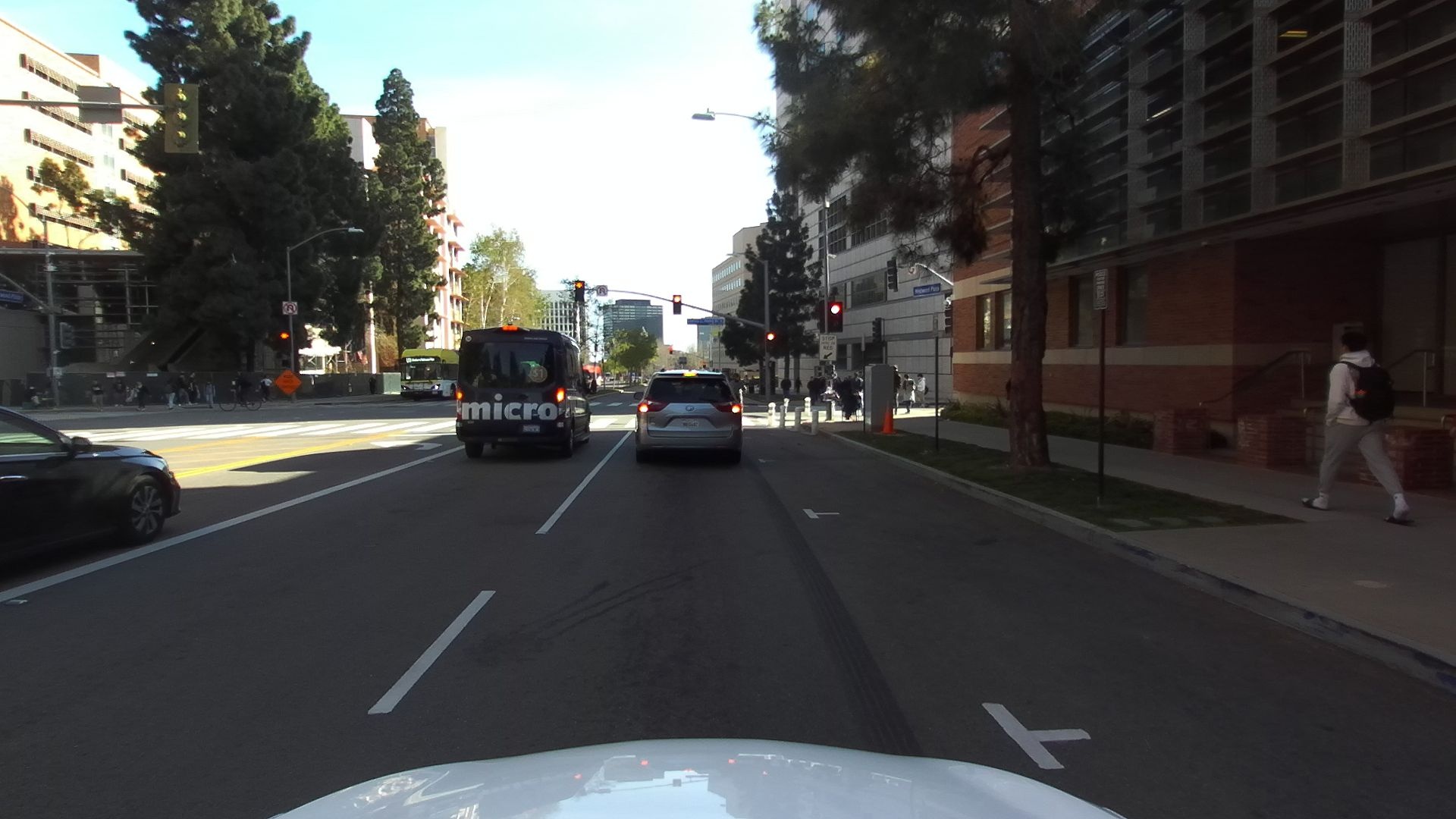}}
        \caption{Left image}
        \label{fig:stereo-a}
    \end{subfigure}
    \begin{subfigure}[c]{0.49\linewidth}
        \centering{\includegraphics[width=1\linewidth]{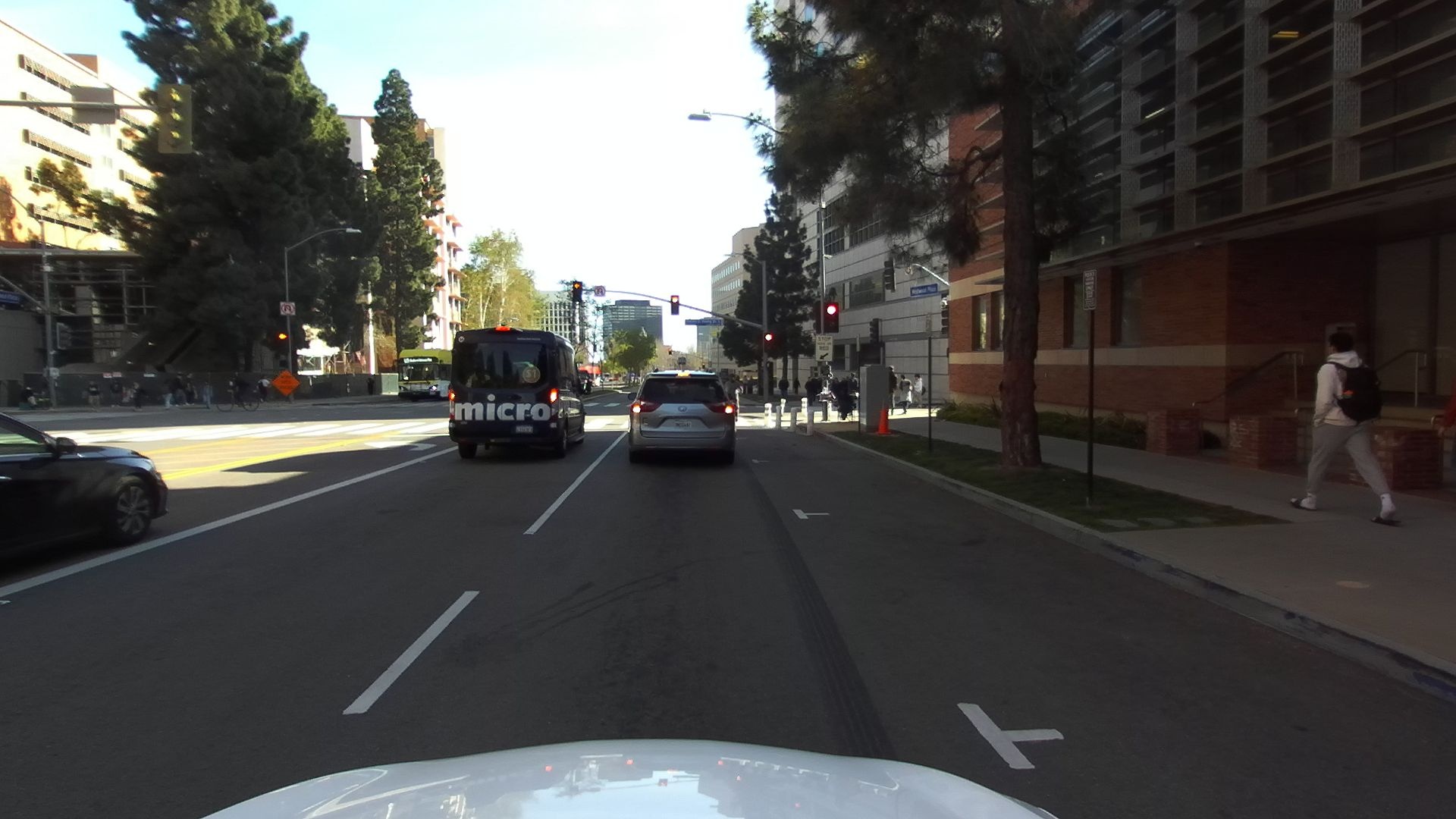}}
        \caption{Right image}
        \label{fig:stereo-b}
    \end{subfigure}
    \caption{Stereo Camera Image Pairs: (a) Image from the left camera. (b) Image from the right camera. }
    \label{fig:stereo}
    \vspace{-6mm}
\end{figure*}

\begin{figure*}[]
\centering
    \begin{subfigure}[c]{0.49\linewidth}
        \centering{\includegraphics[width=1\linewidth]{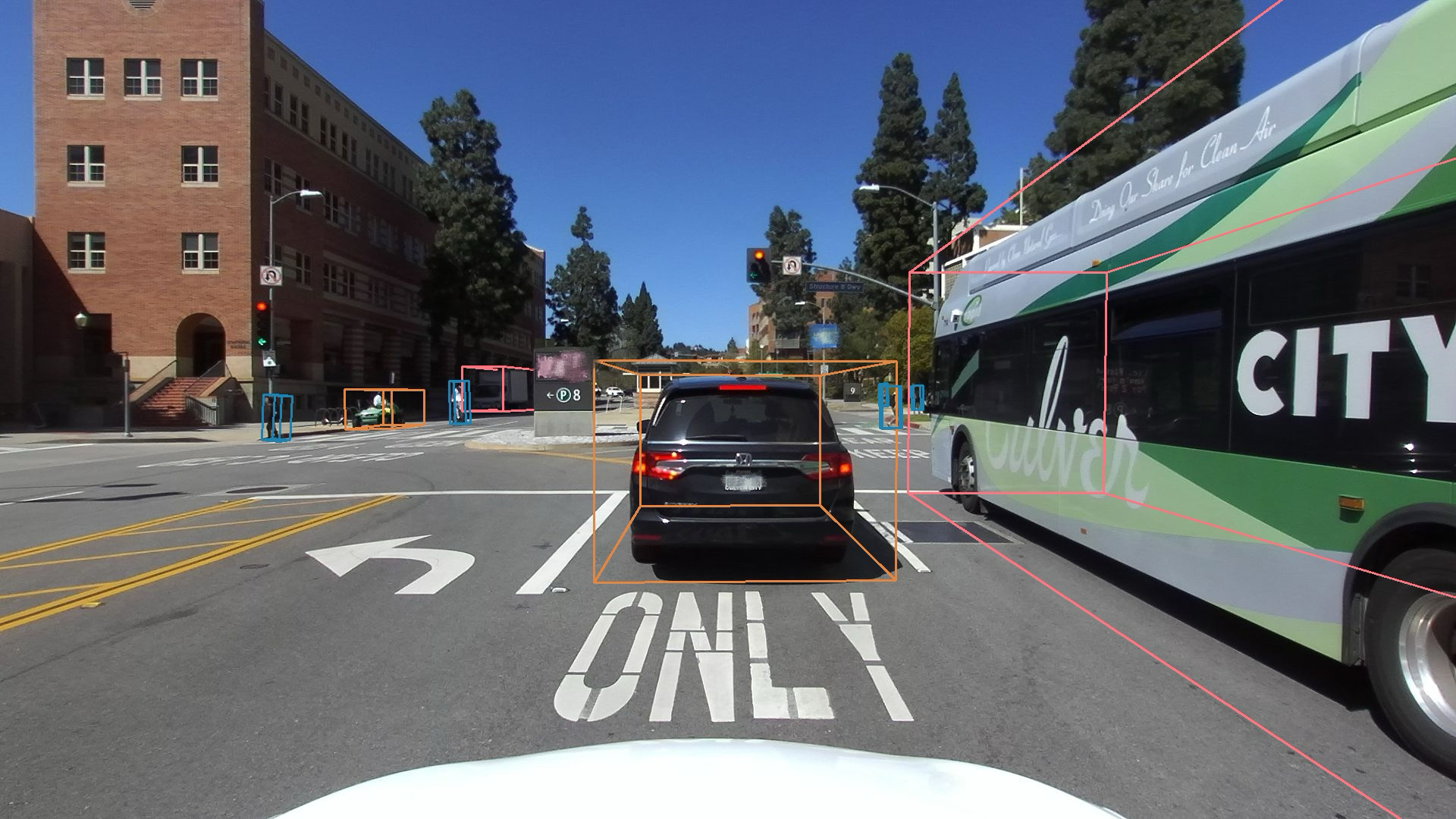}}
        \caption{Front camera}
        \label{fig:bev_camera-a}
    \end{subfigure}
    \begin{subfigure}[c]{0.49\linewidth}
        \centering{\includegraphics[width=1\linewidth]{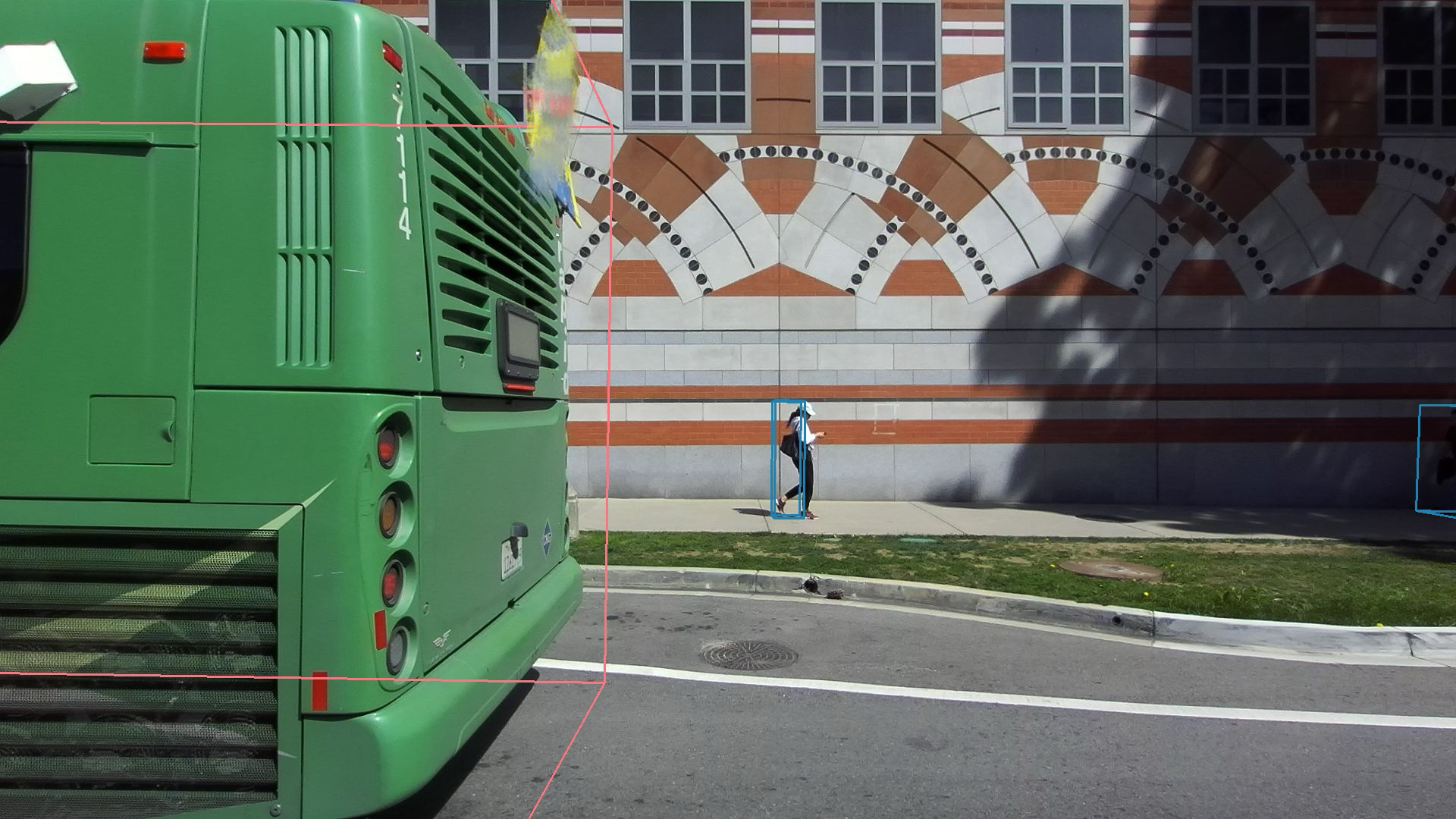}}
        \caption{Right camera}
        \label{fig:bev_camera-b}
    \end{subfigure}
    \begin{subfigure}[c]{0.49\linewidth}
        \centering{\includegraphics[width=1\linewidth]{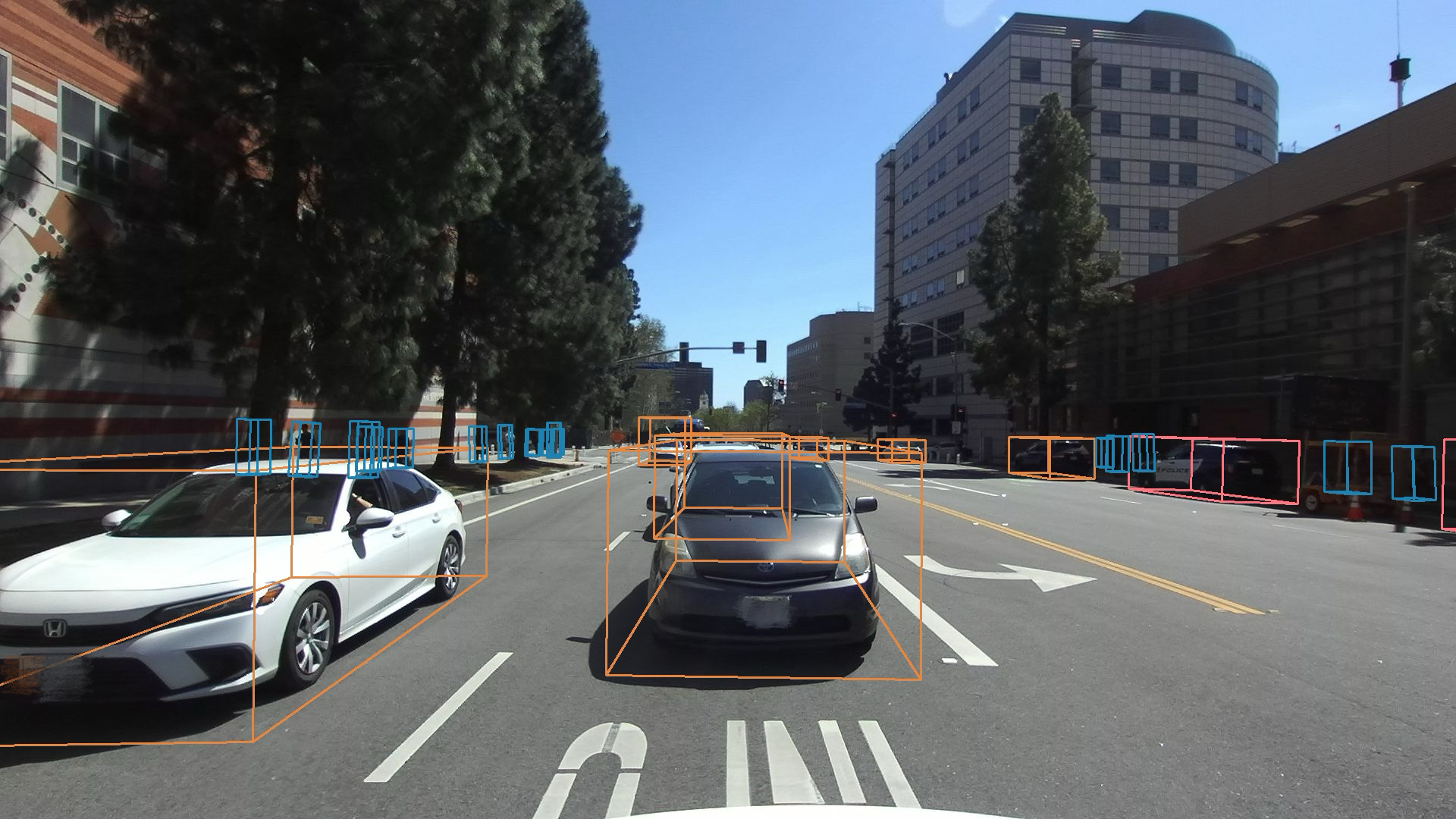}}
        \caption{Rear camera}
        \label{fig:bev_camera-c}
    \end{subfigure}
    \begin{subfigure}[c]{0.49\linewidth}
        \centering{\includegraphics[width=1\linewidth]{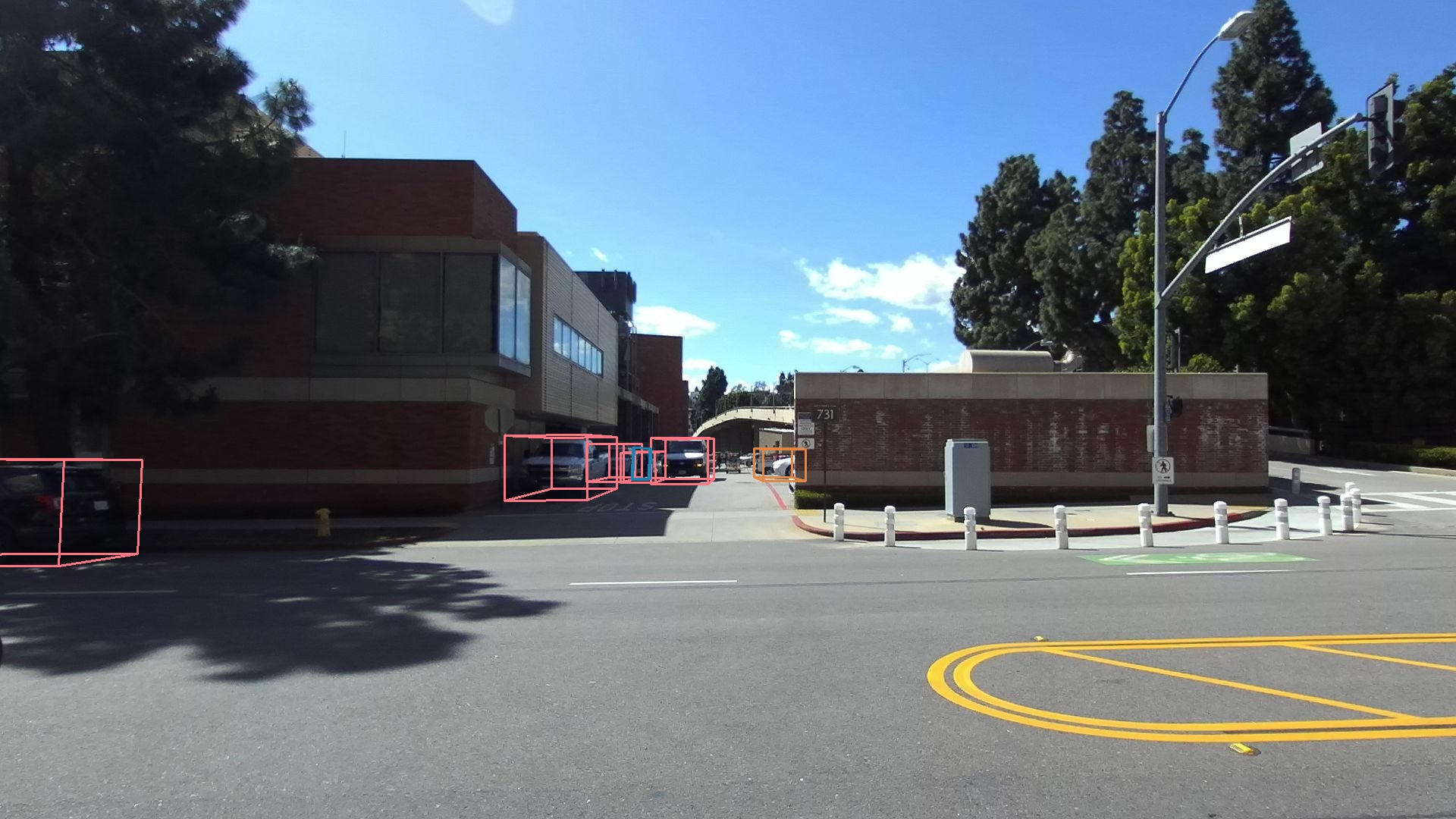}}
        \caption{Left camera}
        \label{fig:bev_camera-d}
    \end{subfigure}
    \caption{Visualization of four stereo cameras of the vehicle. (a)-(d) plot the images from the front, right, rear, and left cameras respectively. 3D bounding boxes are projected onto a 2D image plane and are colored differently as per category. Best viewed in color with zoom for enhanced detail. }
    \label{fig:bev_camera}
    \vspace{-6mm}
\end{figure*}

V2X-Real is collected in two types of scenarios \ie V2X smart intersection and V2V corridors. Visualizations of the smart intersection are presented in Fig. \textcolor{red}{1-3} of the main paper. In this supplementary material, additional visualizations are provided for the V2V corridors (Fig.~\ref{fig:v2v_corridor}). Each figure displays three images: the top image showcases the aggregated LiDAR point clouds from both vehicles, while the bottom images present views from the associated front-view cameras. Further, Fig.~\ref{fig:stereo} illustrates the stereo image pairs from the front camera. Notably, each vehicle is equipped with four stereo cameras, yielding eight images per frame, as illustrated in Fig.~\ref{fig:bev_camera}. These cameras, each offering a 120-degree horizontal field of view (FoV), are mounted in four perpendicular directions to achieve a complete 360-degree FoV around the vehicle. In Fig.~\ref{fig:bev_camera}, ground-truth 3D bounding boxes are projected onto the image plane via the intrinsics and extrinsics and each category \ie, \textcolor{blue}{pedestrian}, \textcolor{orange}{car}, and \textcolor{pink}{truck}, are colored differently.

\begin{figure}[]
\centering
\footnotesize
\def\xwidth{0.4}
\def\yheight{0.2}
\def\xem{-2pt}
\def\im_shift{0.1\textwidth}
\setlength{\tabcolsep}{0.5pt}
\begin{tabular}{cccc}
 & V2X scene & V2V corridor scene\\
 \multirow[t]{1}{*}[\im_shift]{\begin{sideways} No Fusion \end{sideways}} &
\includegraphics[ width=\xwidth\linewidth]{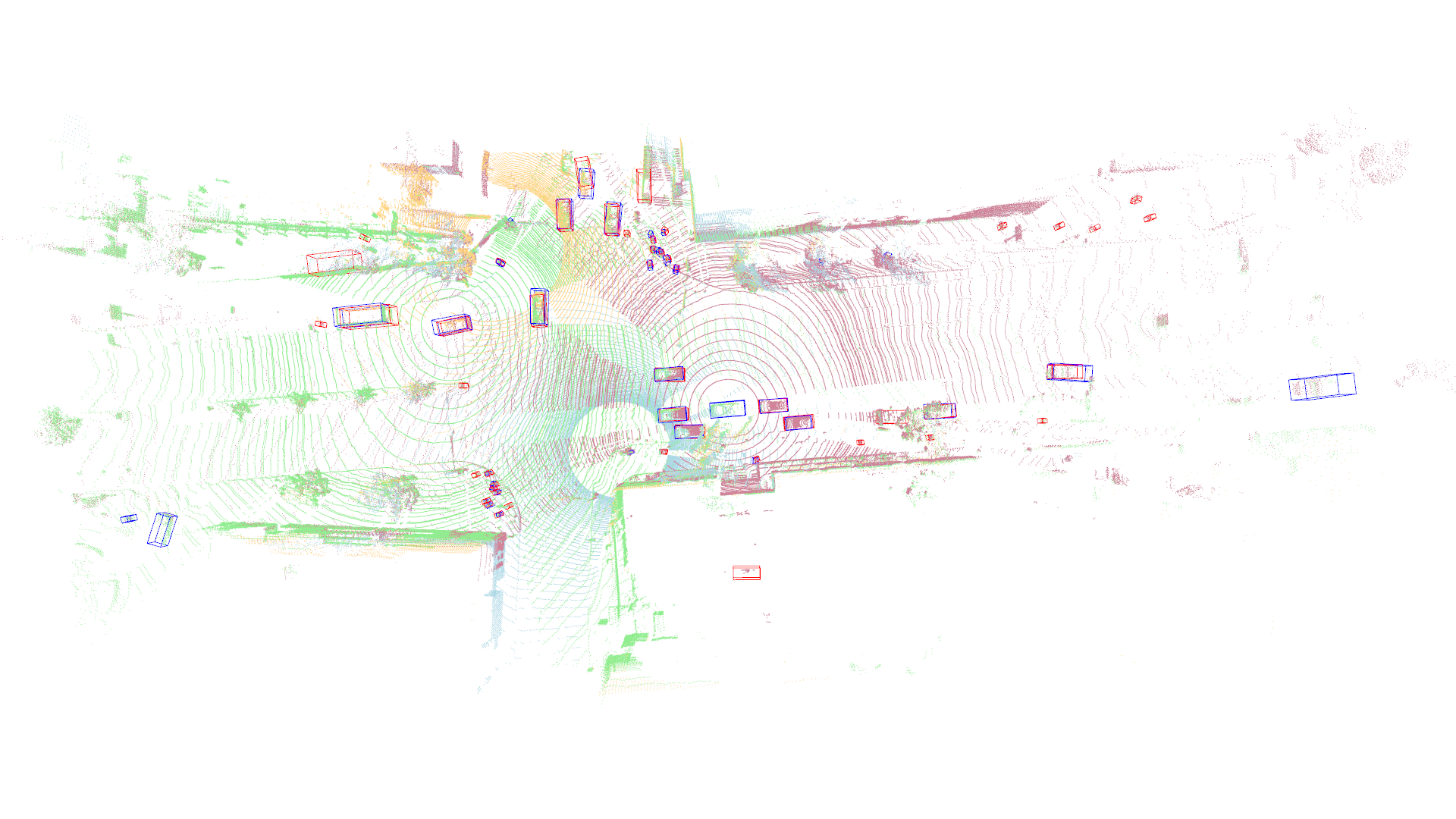}
& \includegraphics[ width=\xwidth\linewidth]{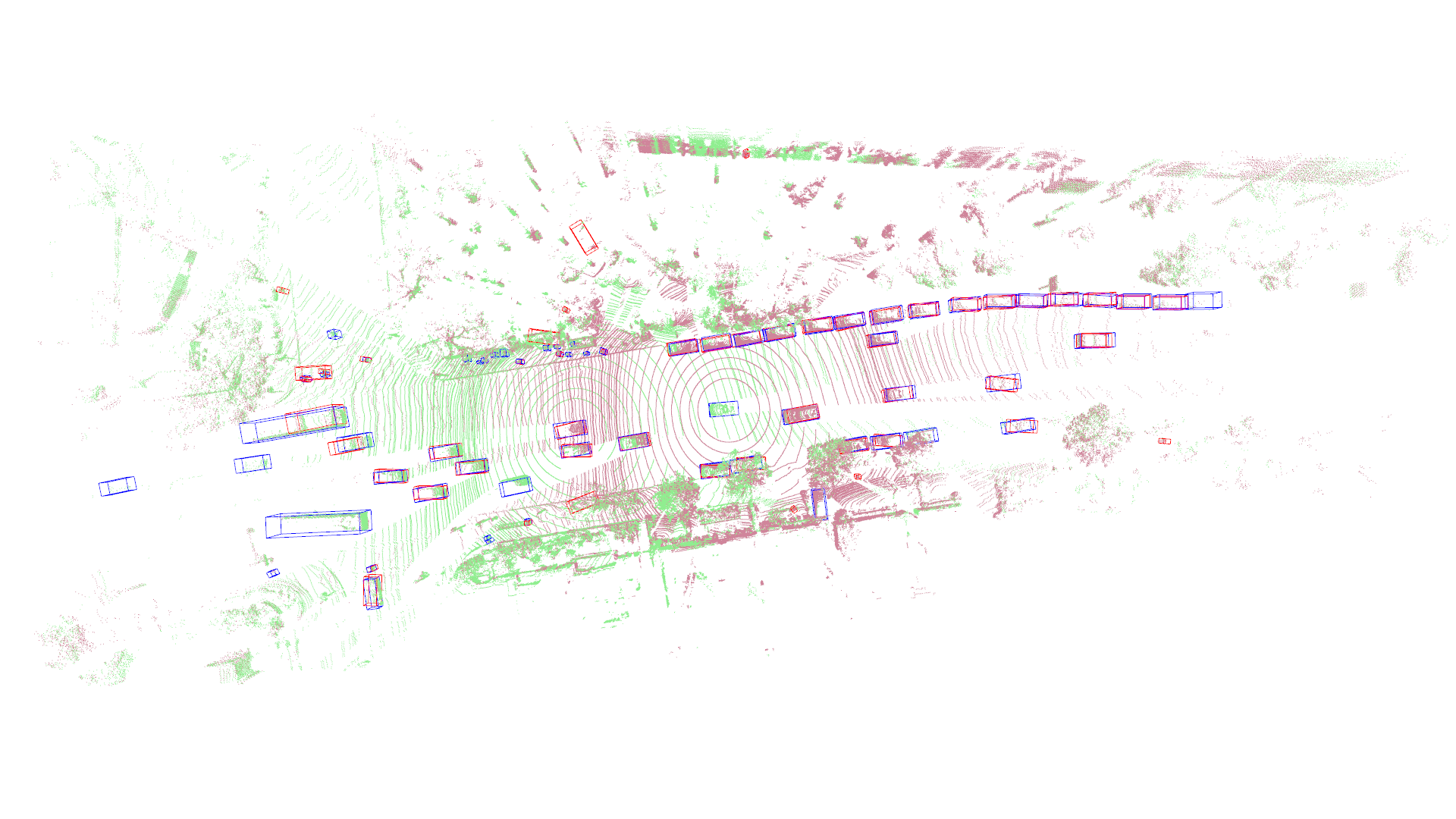}\\

\multirow[t]{1}{*}[\im_shift]{\begin{sideways}  Late Fusion  \end{sideways}}  &
 \includegraphics[ width=\xwidth\linewidth]{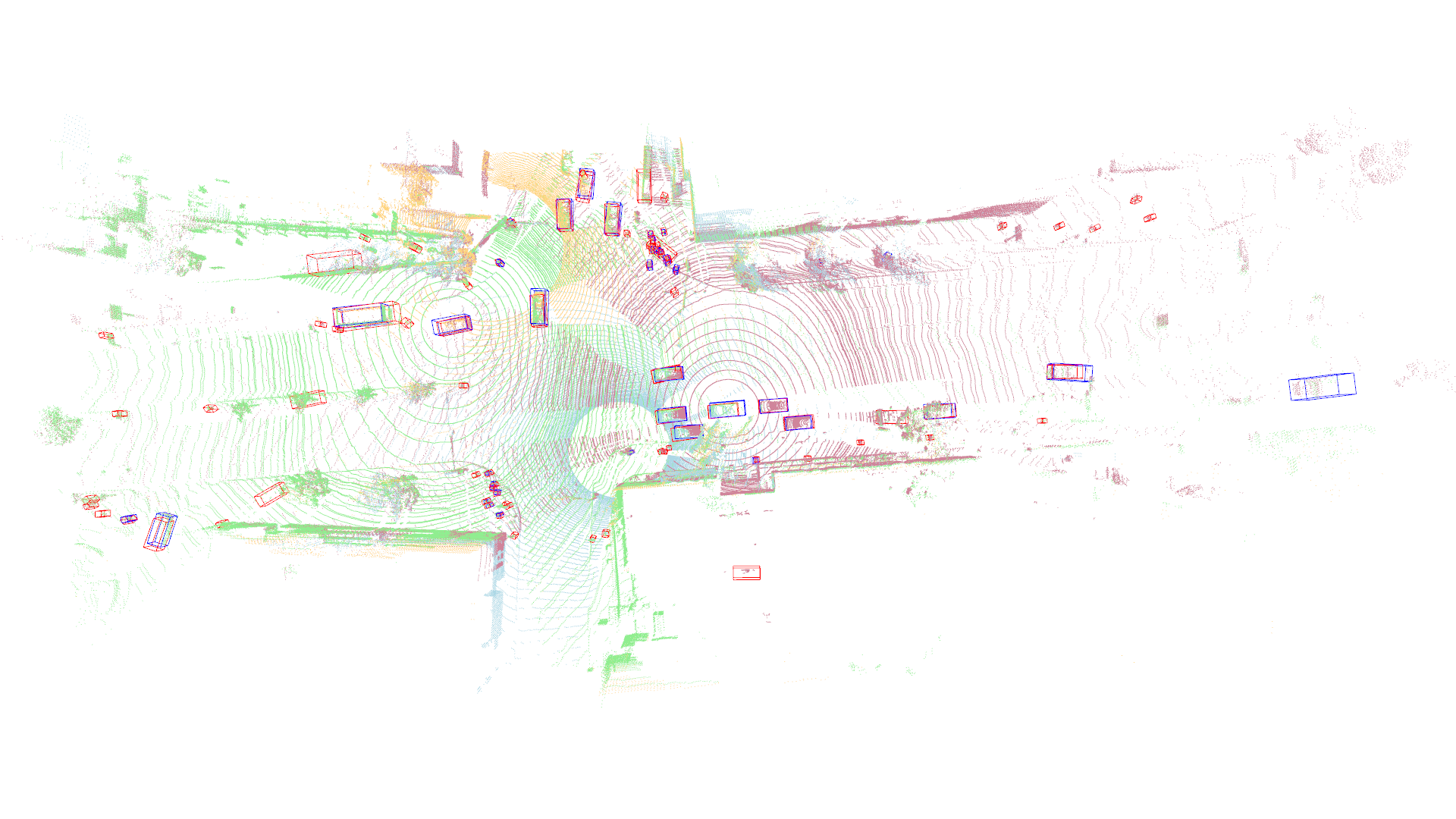}
& \includegraphics[ width=\xwidth\linewidth]{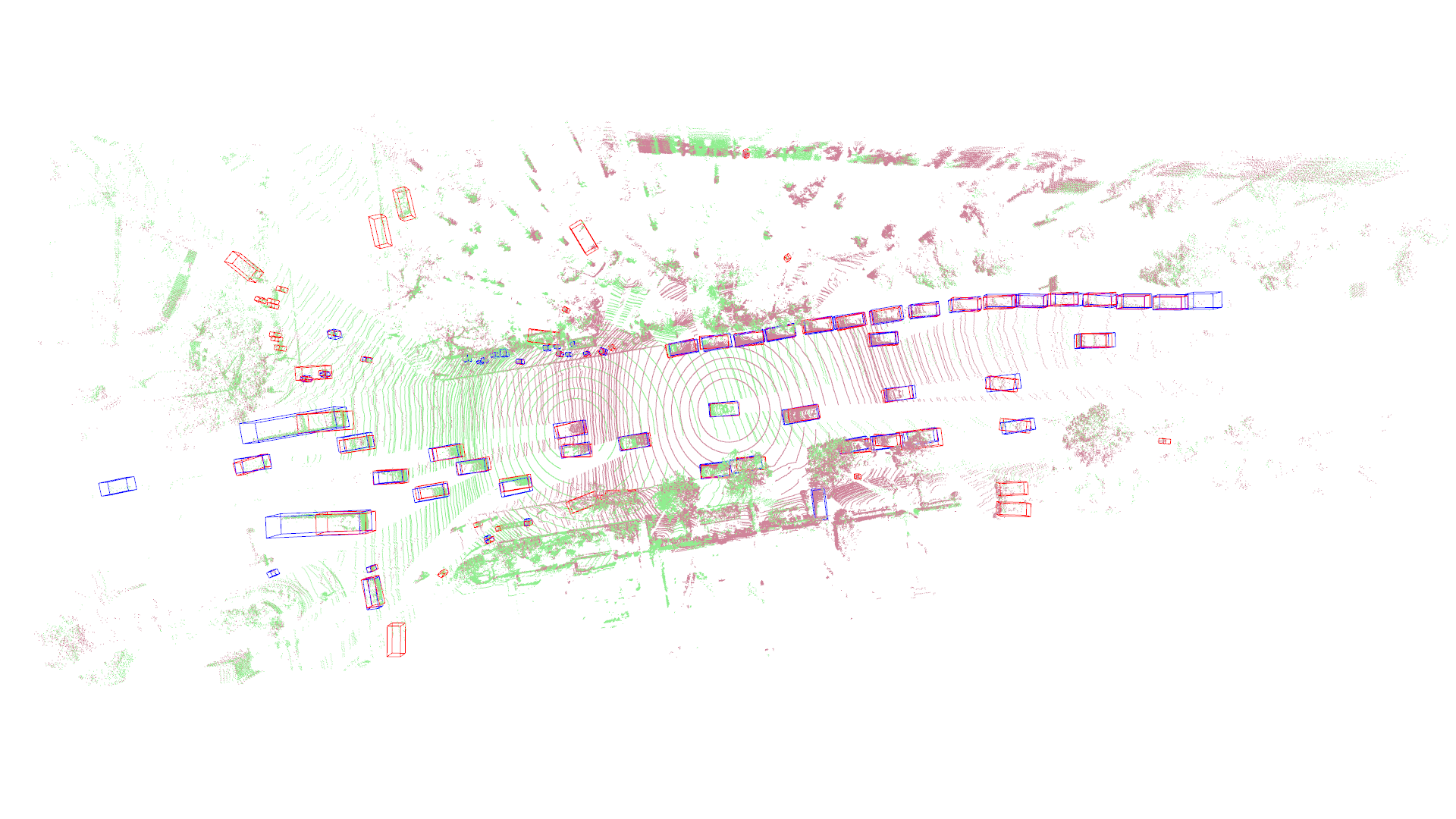}\\

\multirow[t]{1}{*}[\im_shift]{\begin{sideways}  Early Fusion  \end{sideways}}  &
 \includegraphics[ width=\xwidth\linewidth]{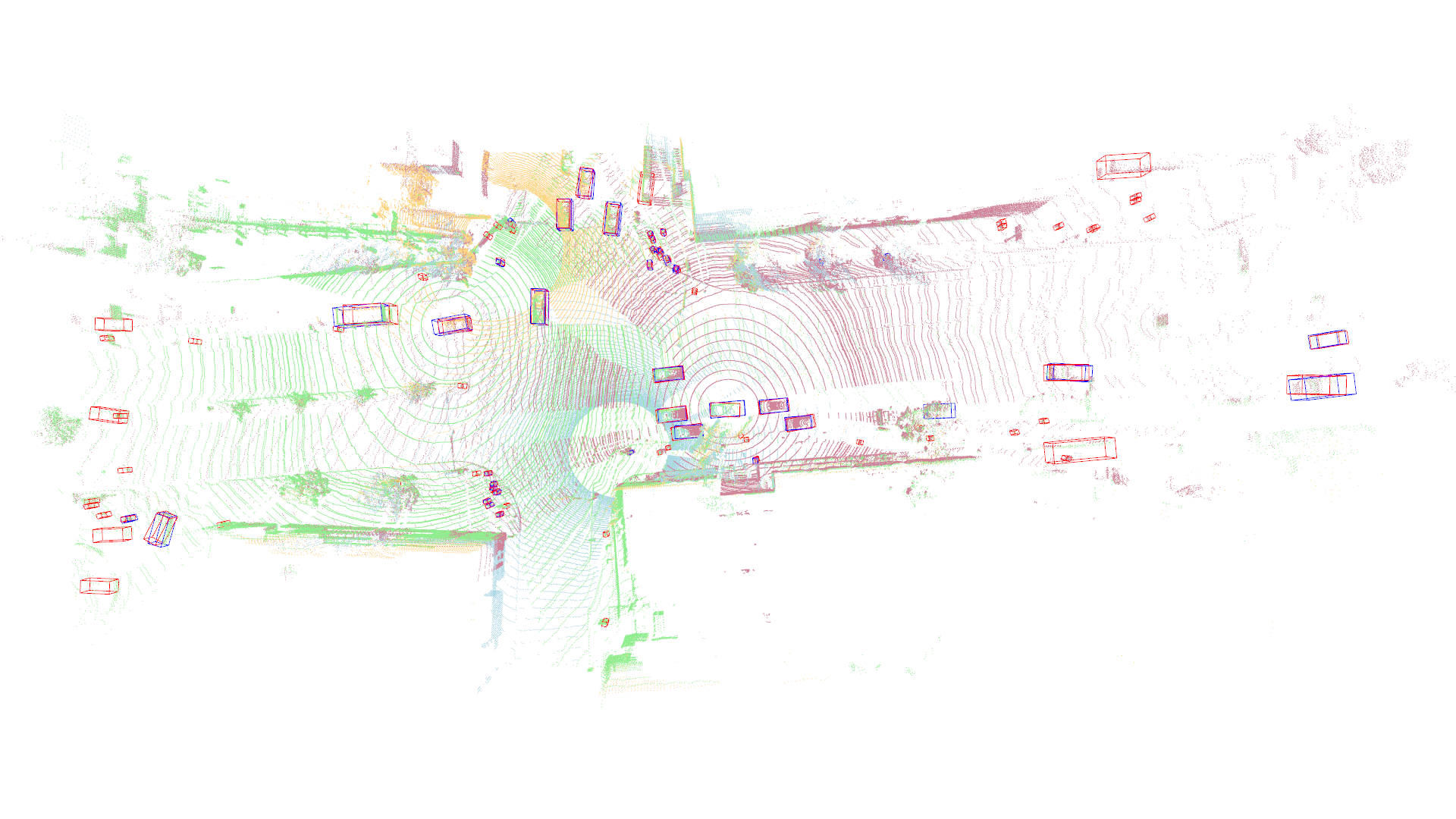}
& \includegraphics[ width=\xwidth\linewidth]{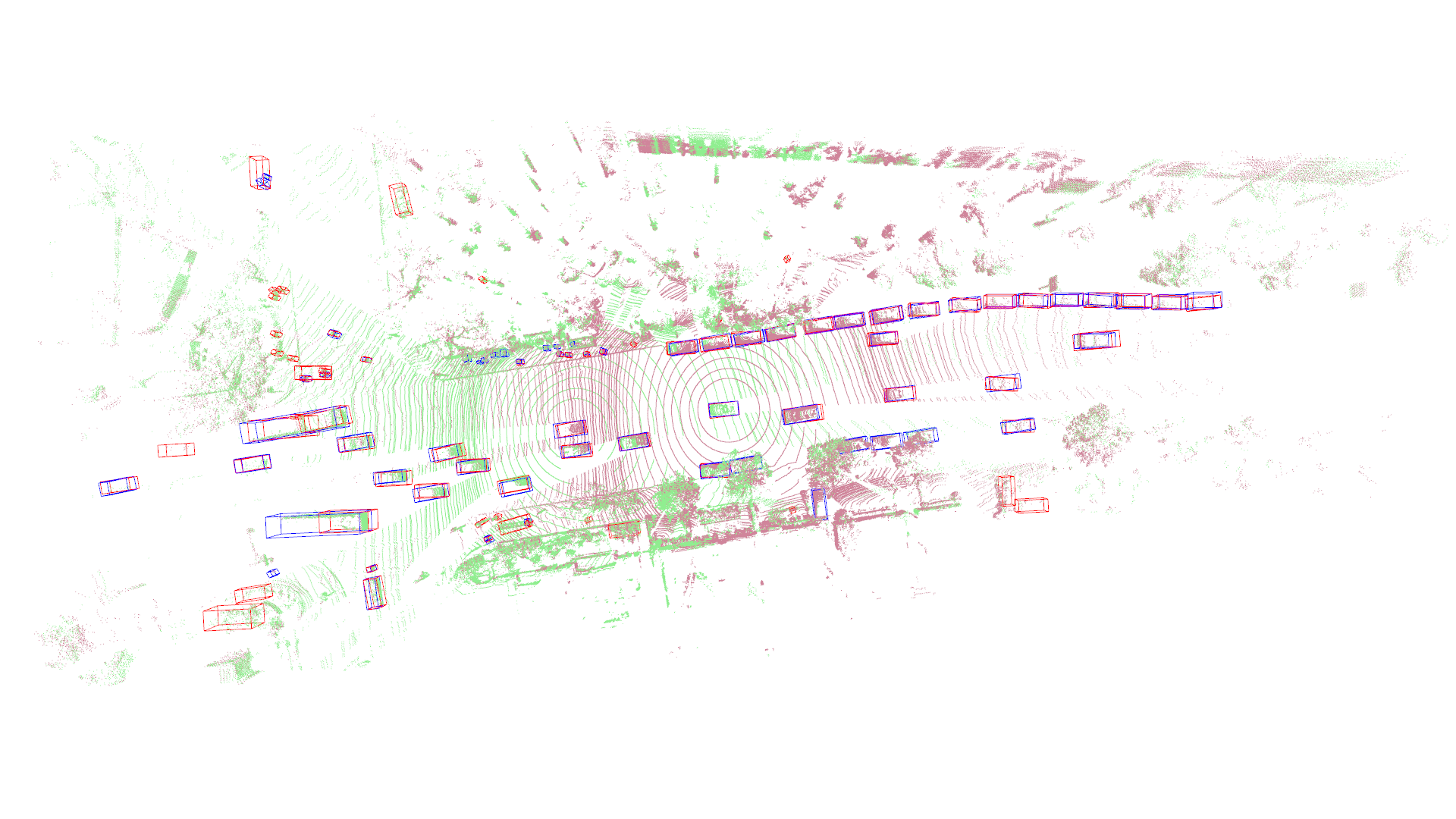}\\

\multirow[t]{1}{*}[\im_shift]{\begin{sideways}  F-Cooper~\cite{chen2019f}  \end{sideways}}  &
 \includegraphics[ width=\xwidth\linewidth]{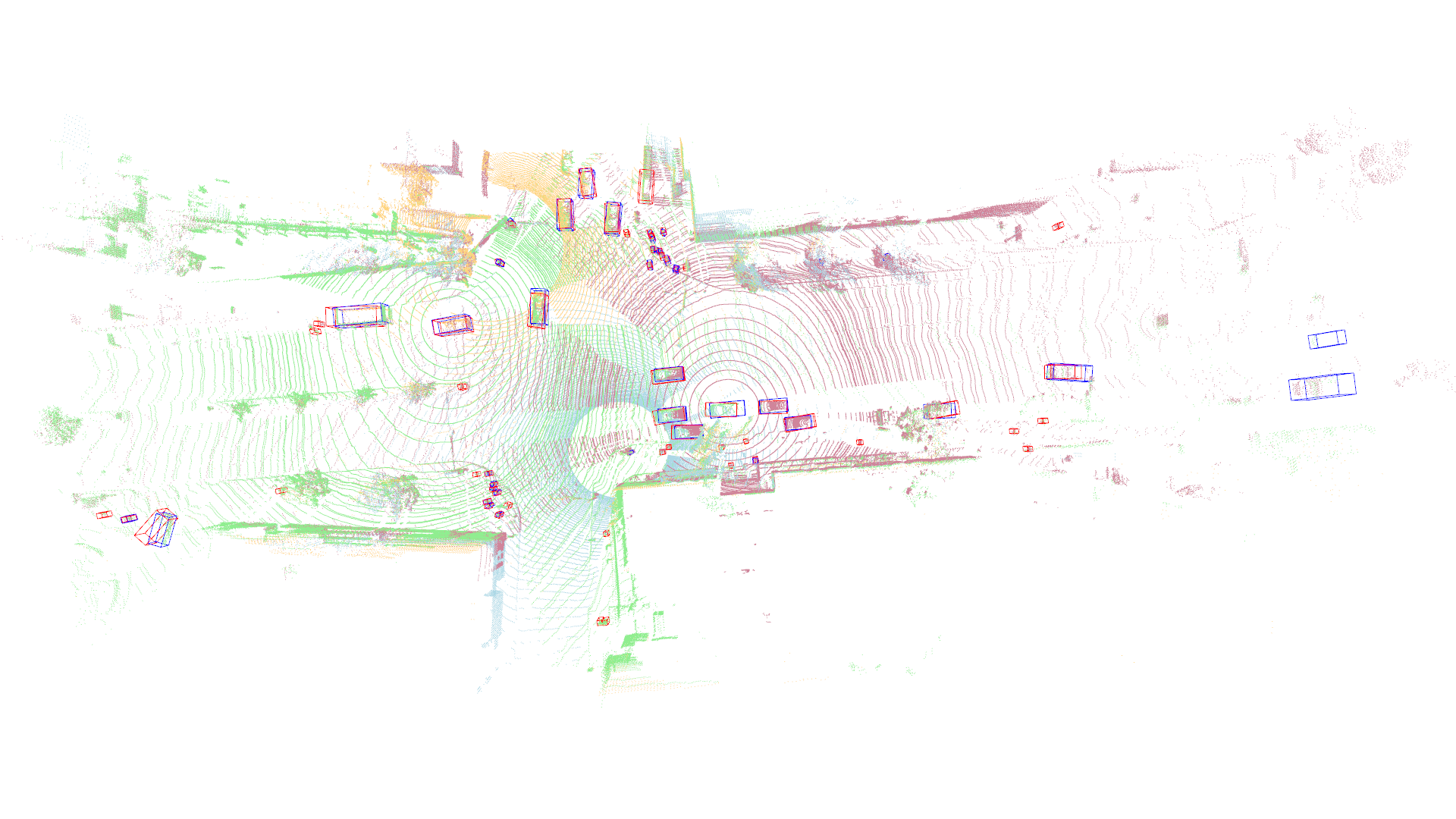}
& \includegraphics[ width=\xwidth\linewidth]{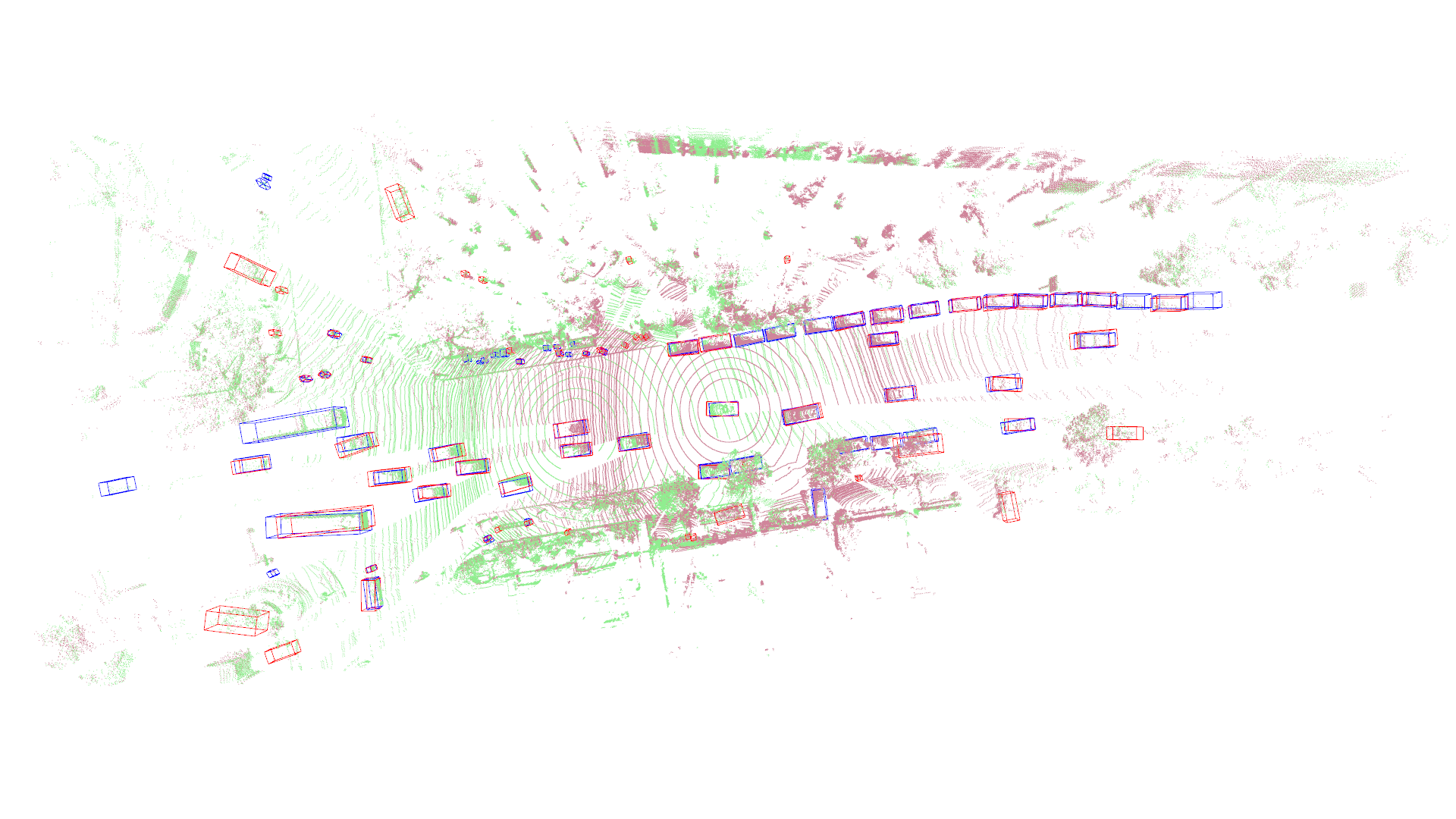}\\
\multirow[t]{1}{*}[\im_shift]{\begin{sideways}  AttFuse~\cite{xu2022opv2v} \end{sideways}} &
\includegraphics[width=\xwidth\linewidth]{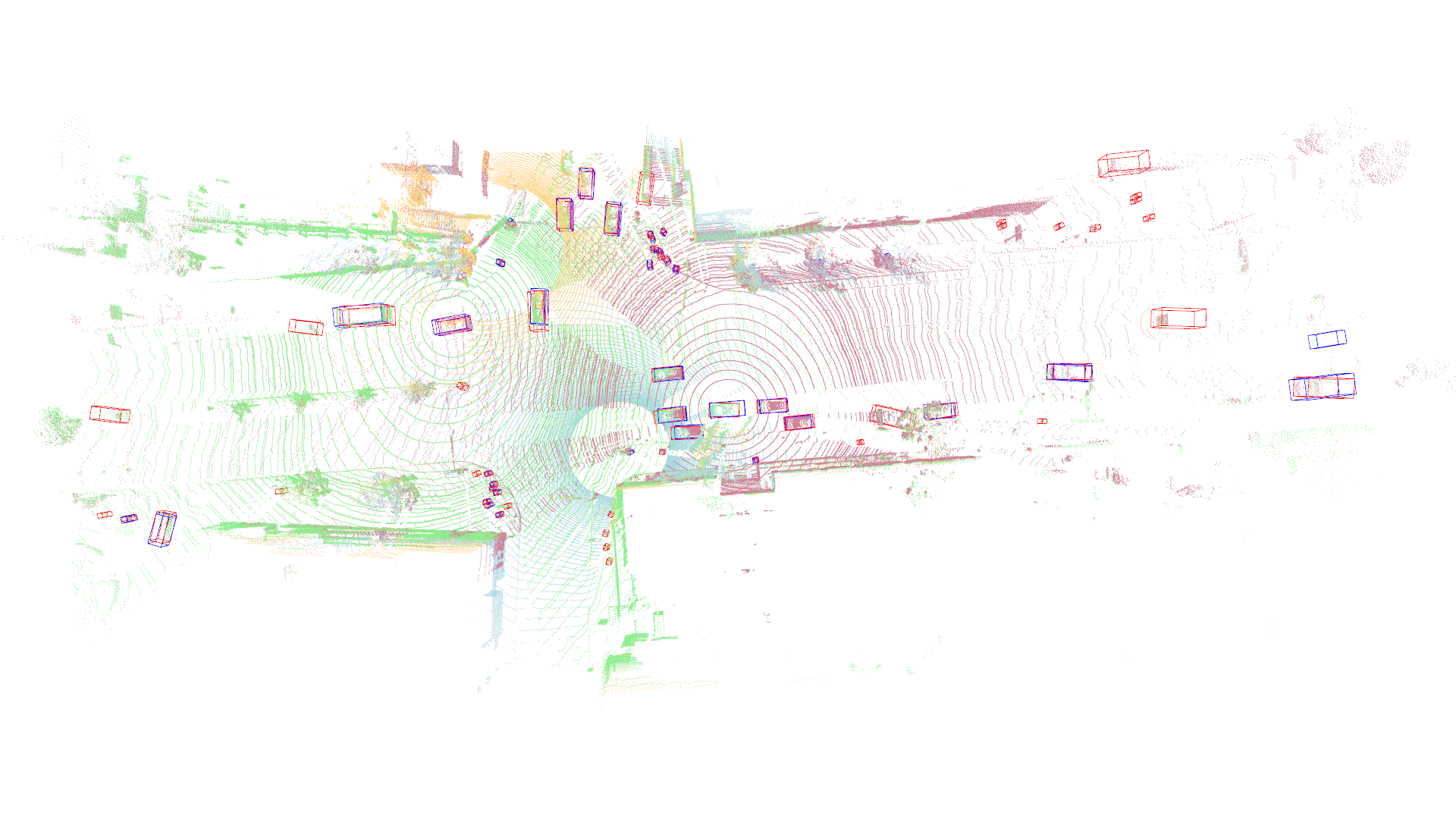}
& \includegraphics[width=\xwidth\linewidth]{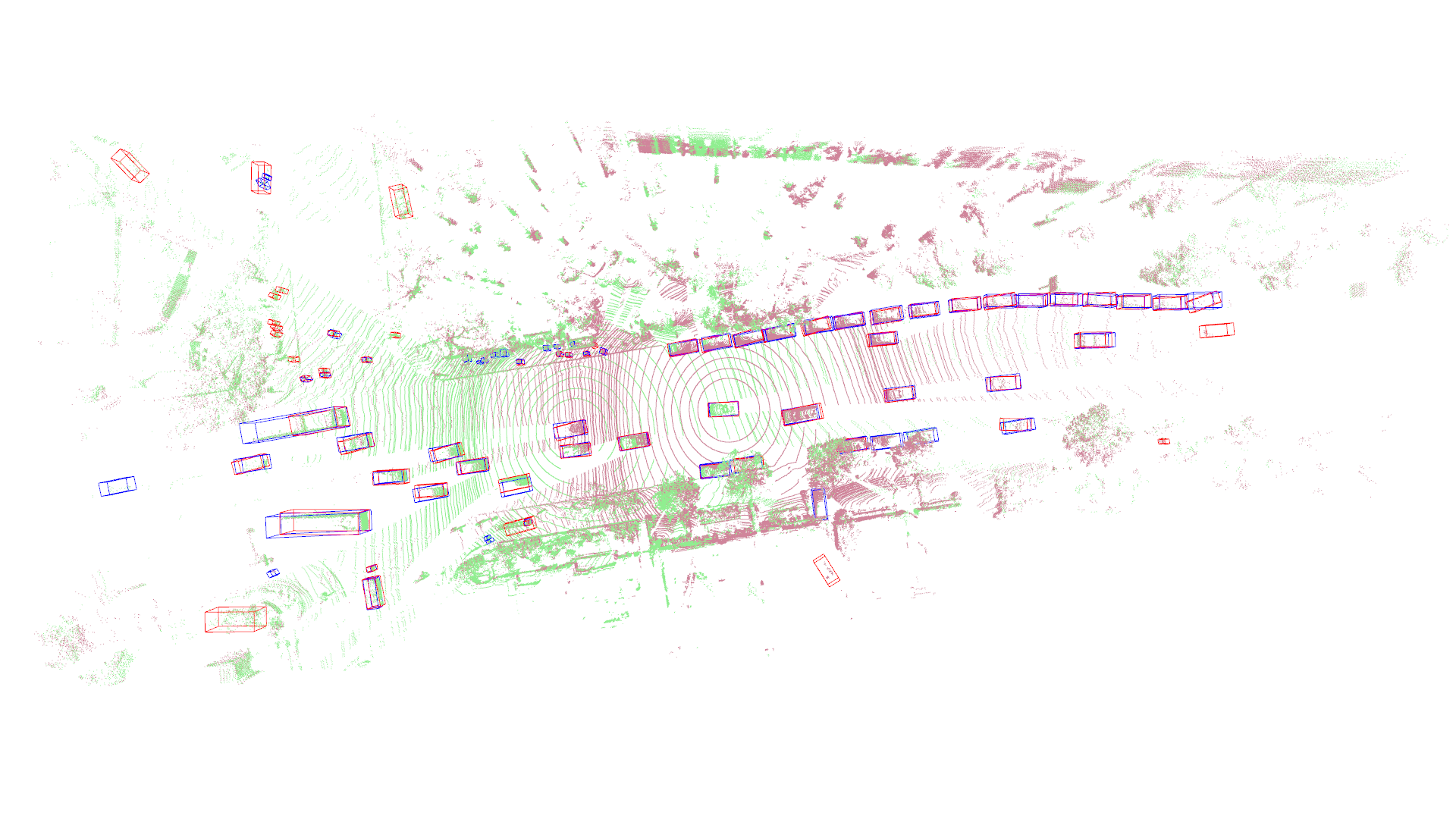}\\
\multirow[t]{1}{*}[\im_shift]{\begin{sideways}  V2X-ViT~\cite{xu2022v2x} \end{sideways}} &
 \includegraphics[ width=\xwidth\linewidth]{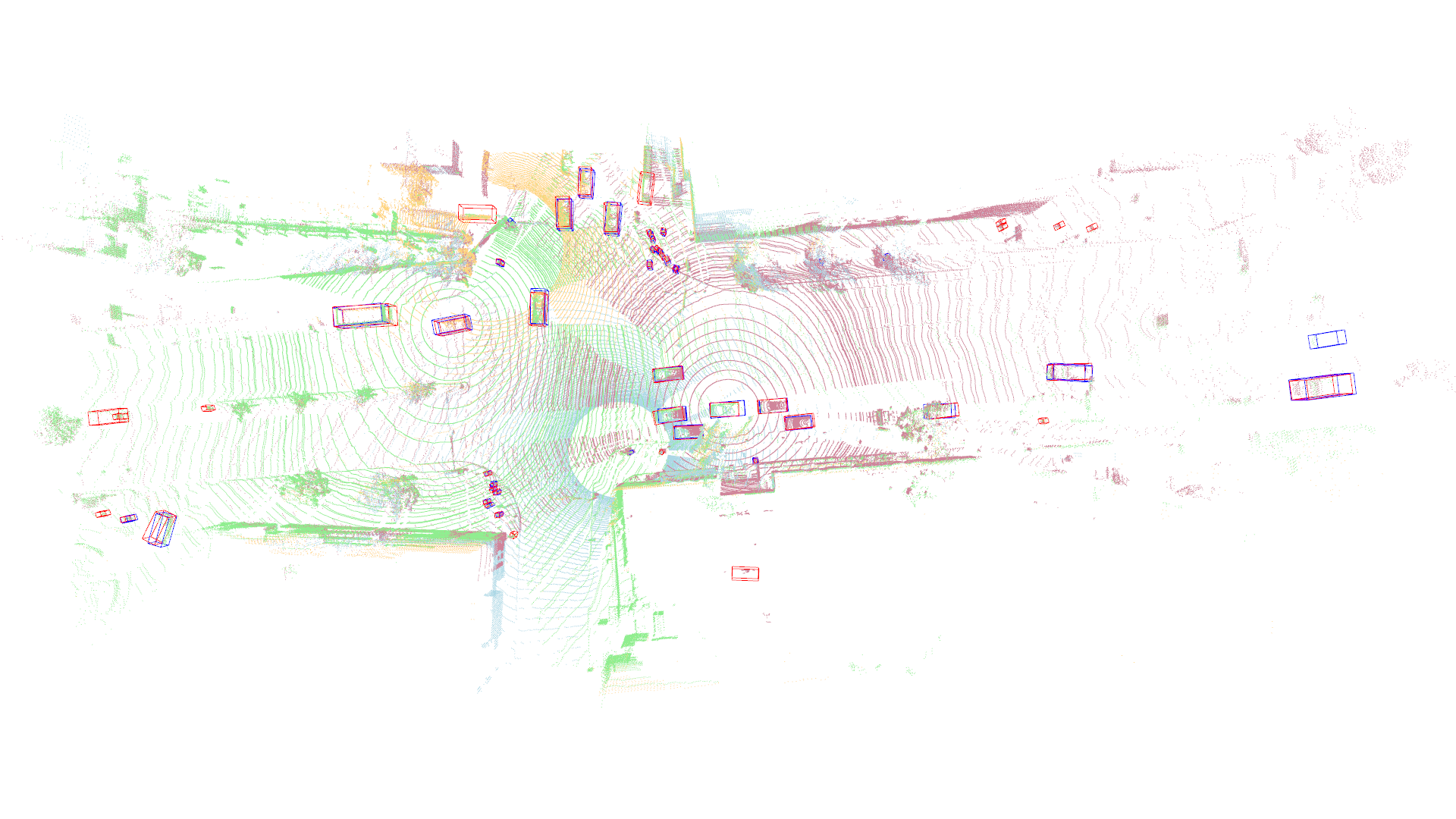}
& \includegraphics[ width=\xwidth\linewidth]{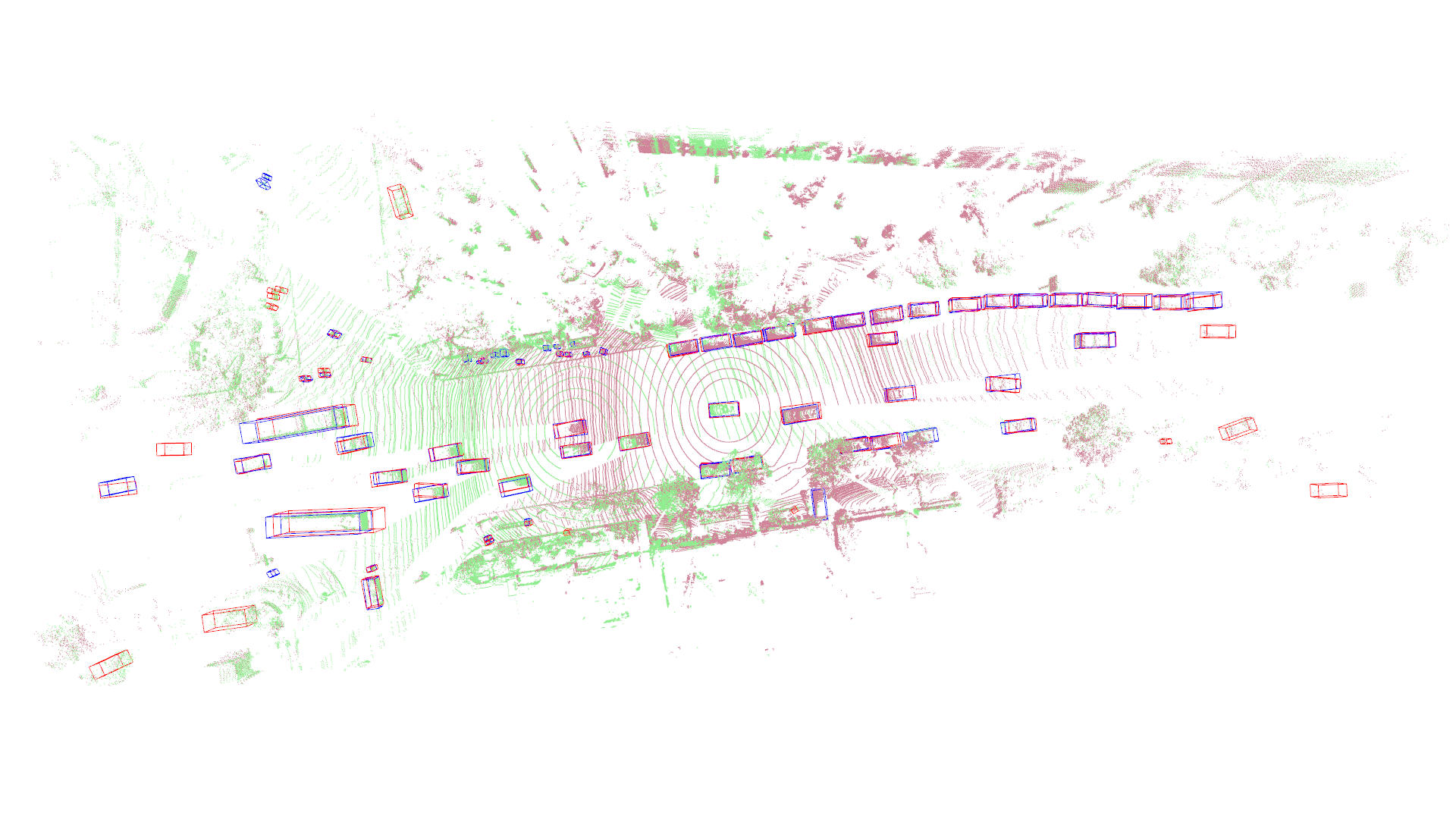}
\end{tabular}
\vspace{-3mm}
\caption{\textbf{Qualitative visualization of cooperative perception in a V2X smart intersection scene and a V2V corridor scene.} \textcolor{blue}{Blue} and \textcolor{red}{red} 3D bounding boxes correspond to the ground-truth and detection outputs, respectively.}
\label{fig:sup-qualitive3}
\end{figure}

\section{Model Details}
\label{sec:model_details}
In this section, we detail the model configurations for the LiDAR-based cooperative perception (Sec.~\ref{sec:lidar-coop}) and Camera-based cooperative perception (Sec.~\ref{sec:camera-coop}).
\subsection{LiDAR-based Cooperative Perception}
\label{sec:lidar-coop}
\noindent\textbf{LiDAR backbone: }We adopt PointPillar~\cite{lang2019pointpillars} as the LiDAR backbone with a voxel size of 0.4 meters in the x and y directions. Following~\cite{xu2022v2x, xu2022opv2v, xiang2023hm}, each agent first project its LiDAR to the ego agent's coordinate frame. Then, the raw points are converted into pillar features and scattered into a 2D Bird's Eye View (BEV) feature map, which is shared with neighboring connected agents. In this work, the number of points per voxel is set to 32 while the maximum voxel number is 64000.  \\
\noindent\textbf{Fusion method: }The ego agent receives all the collaborators' features and produces refined feature representation via fusing all the received feature maps and its own extracted features. We provide benchmarks for three intermediate fusion methods \ie, F-Cooper~\cite{chen2019f}, AttFuse~\cite{xu2022opv2v}, V2X-ViT~\cite{xu2022v2x}. F-Cooper leverages max-pooling to aggregate features. AttFuse adopts per-location attention for feature refinement. V2X-ViT employs multi-scale window attention and heterogeneous self-attention to jointly learn the inter-agent and intra-agent interactions. \\
\noindent\textbf{Head: }We adopt two 1x1 convolution layers for box regression and classification.  The regression branch outputs $(x,y,z,w,h,l,\theta)$, denoting the position $(x,y, z)$, box dimensions $(w,h,l)$ and yaw angle $(\theta)$ with respect to the predefined anchor box of each category. For the classification branch, the head outputs the confidence score of each category \ie, pedestrian, car, truck. To capture the dimension prior of various types of objects, we define a set of fixed-size anchor boxes for each category as per its box size distribution and the network only learns the displacement with respect to these predefined anchor boxes. In this way, the model can efficiently identify multi-scale objects (Fig.~\ref{fig:sup-qualitive3}).

\subsection{Camera-based Cooperative Perception}
\label{sec:camera-coop}
For saving the computation, only the left camera from each stereo camera pair is used, leading to a total of 4 images (Fig.~\ref{fig:bev_camera}) per vehicle. To extract salient features from 2D images,  we employ an architecture similar to BEVFormer~\cite{li2022bevformer} with no temporal information. Following~\cite{xiang2023hm}, for faster running speed, we adopt ResNet50 to extract 2D image features and leverage anchor-based head to generate the final detection outputs. In camera-based cooperative perception, we provide benchmarks for No Fusion and Late Fusion methods.\\ 

\noindent\textbf{No Fusion: }This approach utilizes solely the camera data from a single agent to generate 3D bounding boxes, serving as the baseline that operates without any collaborative input.\\

\noindent\textbf{Late Fusion: }In this method, each agent independently generates 3D bounding box proposals using its own camera data, along with associated confidence scores. These proposals and associated confidence scores are then transmitted to the ego agent, which employs non-maximum suppression (NMS) to produce consistent results from the received proposals.\\

The evaluation for the camera perception is conducted in the range of 50 meters in the x and y directions of the ego coordinate frame. 
\section{More experimental results}
\label{sec:exp}

\subsection{LiDAR-based Cooperative Perception}
Fig.~\ref{fig:sup-qualitive3} offers qualitative visualizations for all the benchmarked methods, confirming the overarching trend observed in the main paper—fusion methods excel over the No Fusion baseline, with Intermediate Fusion methods showing superior precision. Notably, while Late Fusion successfully identifies more true positives compared with No Fusion, it also leads to a rise in false negative predictions. We argue this is due to the loss of context during the late fusion strategy where only the detection results are shared and fused. Consequently, the uncertainty inherent in collaborator messages may be amplified, resulting in an increased rate of false negatives. On the other hand, intermediate fusion methods produce much fewer false positives. We contend that this is due to the end-to-end multi-agent fusion, where the shared/fused features preserve the raw context information and thus can better help refine the ego agent's feature representation, leading to boosted performance.

\begin{table}[]
    \centering
    \caption{Benchmark results of camera-based cooperative perception methods. AP is measured under IoU thresholds of 0.3 and 0.5. }
    \begin{tabular}{c|c|c|c|cc}
    \toprule
    Dataset&Methods&Backbone&Collaboration&AP@0.3&AP@0.5\\
    \midrule
         \multirow{2}{*}{V2X-Real-VC}&No Fusion&BEVFormer~\cite{li2022bevformer}&&14.7&13.5  \\
         &Late Fusion&BEVFormer~\cite{li2022bevformer}&\checkmark&\textbf{16.9}&\textbf{15.0} \\ 
    \midrule
     \multirow{2}{*}{V2X-Real-V2V}&No Fusion&BEVFormer~\cite{li2022bevformer}&&7.3&6.8  \\
         &Late Fusion&BEVFormer~\cite{li2022bevformer}&\checkmark&\textbf{8.2}&\textbf{7.2} \\ 
    \bottomrule
    \end{tabular}
    \label{tab:camera_benchmark}
\end{table}

\subsection{Camera-based Cooperative Perception} In this work, we provide benchmarks for vehicle-side multi-view camera-based cooperative perception and leave infrastructure-side camera solution for future works as the infrastructure camera has a fixed topology with roads, which could serve as a prior for detection task and thus require a dedicated new algorithm for efficient Infrastructure-to-Everything cooperative perception. Tab.~\ref{tab:camera_benchmark} demonstrates the benchmark performance for camera-based cooperative perception on V2X-Real-VC (excluding infrastructure collaborator) and V2X-Real-V2V. The results indicate that Late Fusion outperforms No Fusion baseline for both datasets. Moreover, both models perform higher in the V2X-Real-VC dataset than the V2X-Real-V2V, which contains V2V corridor scenarios. We argue this is due to the fact that the V2V corridors contain more diverse scenarios and roads with high slope changes, posing challenges for BEV-based camera solutions in accurately determining 3D poses. We encourage further exploration into these important yet under-explored areas by the research community.

%
%
\bibliographystyle{splncs04}
\bibliography{main}
\end{document}